\documentclass{ieeetj}
\usepackage{cite}
\usepackage{amsmath,amssymb,amsfonts}
\usepackage{algorithmic}
\usepackage{graphicx,color}
\usepackage{textcomp}
\usepackage{xcolor}
\usepackage{hyperref}
\hypersetup{hidelinks=true}
\usepackage{array}
\usepackage{stfloats}
\usepackage{url}
\usepackage{subcaption}
\usepackage{verbatim}
\usepackage{graphicx}
\usepackage{rotating}
\usepackage{tabularray}
\usepackage{tabularx}
\usepackage{graphicx}
\usepackage{ragged2e}
\usepackage{booktabs}
\usepackage{multirow}
\usepackage{balance}
\usepackage[T1]{fontenc}
\usepackage{algorithm,algorithmic}
\def\BibTeX{{\rm B\kern-.05em{\sc i\kern-.025em b}\kern-.08em
    T\kern-.1667em\lower.7ex\hbox{E}\kern-.125emX}}
\AtBeginDocument{\definecolor{tmlcncolor}{cmyk}{0.93,0.59,0.15,0.02}\definecolor{NavyBlue}{RGB}{0,86,125}}

\def\OJlogo{\vspace{-4pt} }
\def\seclogo{\vspace{10pt} }

\def\authorrefmark#1{\ensuremath{^{\textbf{#1}}}}

\begin{document}
\receiveddate{10 Oct, 2025}
\reviseddate{10 Oct, 2026}
\doiinfo{XXXX.2026.1234567}

\markboth{}{Shah {et al.} Multi-Scale Spectral Attention Module-based Hyperspectral Segmentation in Autonomous Driving Scenarios}

\title{Multi-Scale Spectral Attention Module-based Hyperspectral Segmentation in Autonomous Driving Scenarios}

\author{
Imad Ali Shah\authorrefmark{1,2},
Jiarong Li\authorrefmark{1,2},
Tim Brophy\authorrefmark{1,2},
Enda Ward\authorrefmark{3},
Martin Glavin\authorrefmark{1,2},
Edward Jones\authorrefmark{1,2} and 
Brian Deegan\authorrefmark{1,2}
}
\affil{School of Engineering, University of Galway, Ireland}
\affil{Ryan Institute, University of Galway, Ireland}
\affil{Valeo Vision Systems, Tuam, Ireland}
\corresp{CORRESPONDING AUTHOR: Imad Ali Shah (e-mail: i.shah2@universityofgalway.ie).}
\authornote{This work was produced by the Connaught Automotive Research (CAR) group at the University of Galway and was supported, in part, by Taighde Éireann – Research Ireland grants 13/RC/2094\_P2 and 18/SP/5942, and co-funded under the European Regional Development Fund through the Southern and Eastern Regional Operational Programme to Lero - the Research Ireland Centre for Software (www.lero.ie), and by Valeo Vision Systems.}

\vspace{-1.5em}
\begin{abstract}
Recent advances in autonomous driving (AD) have highlighted the potential of hyperspectral imaging (HSI) for enhanced environmental perception. However, existing semantic segmentation architectures are predominantly designed for RGB and do not fully exploit the rich spectral information available in HSI data. A key limitation of current HSI processing methods is that spectral bands are often processed independently or uniformly, limiting the ability to capture spectral correlations associated with different materials. This paper presents an empirical study of a Multi-Scale Spectral Attention Module (MSAM) integrated into UNet skip connections (UNet-SC) for HSI semantic segmentation. MSAM uses three parallel 1D convolutions with selected kernel sizes between 1 and 11 to investigate whether multi-scale spectral filtering improves segmentation performance under the constraints of small, imbalanced datasets and heterogeneous spectral coverage. The proposed configuration is evaluated across publicly available multi-class urban HSI datasets. Compared with the UNet-SC baseline, UNet-MSAM achieves average improvements of 2.32\% in mIoU and 2.88\% in mF1 across all datasets and backbone depths, while maintaining competitive GPU inference latency and performance relative to established attention mechanisms. The results further indicate that optimal kernel combinations are dataset-specific, with configurations such as \textit{(1;5;11)} and \textit{(3;7;11)} demonstrating strong and consistent performance across multiple datasets. These findings provide empirical evidence that multi-scale spectral feature extraction can improve hyperspectral segmentation performance and establish a foundation for future work on adaptive spectral attention mechanisms for ADAS/AD perception applications. 

\end{abstract}
\vspace{-1.5em}
\vspace{-1.5em}
\maketitle
\section{INTRODUCTION}
\IEEEPARstart{A}{utonomous} driving (AD) systems require robust and reliable perception capabilities to ensure safe navigation in diverse environmental conditions. While RGB-based computer vision has been the cornerstone of Advanced Driver Assistance Systems (ADAS), its fundamental limitations and inability to characterise material properties~\cite{li2025hyperspectral} motivate the exploration of other sensing modalities. Hyperspectral Imaging (HSI) has emerged as a promising complement to conventional RGB cameras, offering tens to hundreds of spectral bands across the electromagnetic spectrum and thereby enabling detailed material characterisation and improved object discrimination~\cite{martinez2019most} in driving scenes. 

\begin{figure}[!t]
\centering
\includegraphics[width=3.25in]{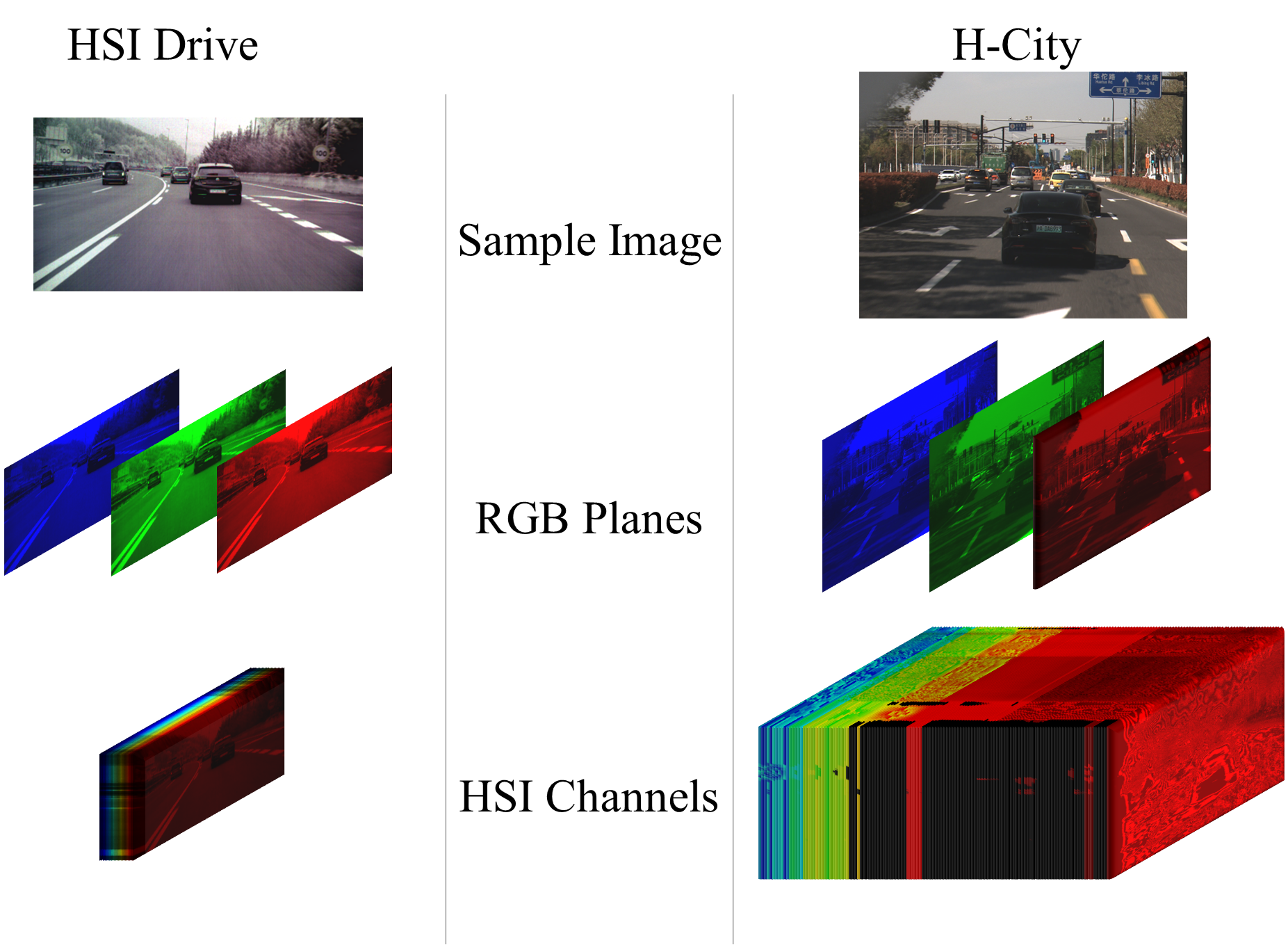}
\caption{Comparison of RGB and HSI: (left) HSI-Drive's pseudo-RGB~\cite{gutierrez2023hsi}  vs (right) H-City’s RGB images~\cite{shen4560035urban} along with their 25 and 128 HSI channels, respectively.}
\label{fig:RGBvsHSI}
\vspace{-4ex}
\end{figure}

Unlike RGB cameras that capture only three spectral bands, HSI data cubes contain a continuous spectral signature for each pixel, as shown in Fig.~\ref{fig:RGBvsHSI} with samples from the HSI-Drive v2 (HSI-Drive)~\cite{gutierrez2023hsi}, and Hyperspectral City v2 (H-City)~\cite{shen4560035urban} datasets.
One of the key potential benefits of HSI in AD is its ability to address metamerism~\cite{huang2021weakly}, a phenomenon where objects with different material compositions appear identical under specific lighting conditions with standard RGB imaging, but they can be distinguished through their unique spectral signatures. This capability has the potential to improve road hazard identification, material differentiation, including water, snow, and ice on the road~\cite{valme2024road}, and pedestrian separability~\cite{li2025hyperspectral}. The integration of HSI into ADAS presents unique challenges distinct from its traditional applications, such as in remote sensing and medicine, where the relatively static image acquisition-based scenarios can accommodate line-scan and slow frame rate HSI cameras. However, AD scenarios require portable snapshot-based HSI cameras capable of real-time video capture~\cite{shah2025hyperspectral}.

Recent technological advances have led to the development of compact and cost-effective snapshot HSI cameras~\cite{gutierrez2022exploring}, enabling the collection of annotated HSI datasets specifically for ADAS scenarios. However, achieving broad spectral and spatial coverage with affordable snapshot cameras remains challenging, particularly while maintaining real-time processing capabilities. This has limited the development of perception systems that leverage the capabilities of HSI and, consequently, the majority of HSI studies in the ADAS/AD have focused on gathering data and presenting initial findings, largely focusing on individual datasets.

Despite the inherent challenges of HSI in the ADAS/AD domain, recent studies have investigated the use of HSI data and have established performance benchmarks, such as Gutiérrez-Zaballa et al.~\cite{gutierrez2022exploring} and Theisen et al.~\cite{theisen2024hs3}, which compared HSI with RGB. Shah et al.~\cite{shah2024hyperspectral} evaluated baseline semantic segmentation models on HSI data, using the available annotated datasets and utilizing deep learning architectures. These studies have demonstrated the potential of HSI while highlighting its unique challenges. However, Shen et al.~\cite{shen4560035urban} indicate that existing deep learning architectures are primarily optimized for RGB, which often does not fully exploit the HSI's rich spectral information. This limitation is mainly due to two key factors: (1) These models are originally designed to extract spatial domain features from RGB images, and the use of pre-trained models (e.g., on ImageNet~\cite{russakovsky2015imagenet}) primarily benefits RGB images. (2) When applied to HSI data, researchers typically reduce the high-dimensional spectral information to fit RGB-oriented models through dimensionality reduction techniques such as principal component analysis and band subselection~\cite{alkhatib2022dimensionality}, potentially leading to loss of discriminative spectral information~\cite{shah2026learnable}. This limitation highlights the need for specialized deep learning architectures that can effectively utilize dense spectral information of HSI.

Existing HSI segmentation approaches in ADAS/AD commonly process spectral channels either independently or through fixed aggregation strategies that do not explicitly model multi-scale spectral dependencies, which limits the ability to distinguish materials exhibiting subtle spectral variations. This motivates the investigation of multi-scale spectral feature extraction mechanisms capable of capturing both short-range and broader spectral correlations. Drawing on findings from hyperspectral remote sensing studies~\cite{sun2024hyperspectral} and prior channel-wise spectral feature extraction approaches for ADAS/AD applications~\cite{shah2024hyperspectral}, multi-scale spectral feature extraction is investigated as a means of improving the utilisation of HSI data for semantic segmentation. The proposed approach employs parallel 1D convolution (conv) kernels to capture spectral dependencies at multiple scales, followed by a fusion-based attention mechanism. The resulting module is integrated into the Skip Connections (SC) variant~\cite{mao2016image} of the UNet~\cite{ronneberger2015u} architecture. Experimental results indicate consistent average performance improvements over the baseline UNet-SC across the evaluated datasets.

The primary contribution of this work is a systematic empirical evaluation of multi-scale spectral attention for HSI semantic segmentation in AD scenarios, rather than proposing a new backbone. The study investigates the effect of multi-scale 1D spectral filtering, evaluates several integration points within UNet, compares the method with established attention methods, and reports dataset-specific kernel behavior under identical training conditions. 

The paper is organized as follows: Section~\ref{sec:Current State} reviews the state of HSI perception for ADAS/AD; Section~\ref{sec:Methodology} presents the MSAM methodology; Section~\ref{sec:Experiments} presents experiments and results; and Section~\ref{sec:Conclusion} concludes the study with contributions and future directions.

\begin{table*}[!t]
\centering
\caption{Summary of publicly available snapshot-based HSI datasets for ADAS/AD scenarios in Urban and Rural Settings. Note: Only the latest versions are included as the specification remains the same for older versions, except for dataset sizes.}
\begin{tabular*}{\textwidth}{@{\extracolsep{\fill}}lcccccc|ccc}
\hline
\textbf{HSI Dataset\textsuperscript{*}} & \textbf{Year} & \textbf{Wavelength (nm)} & \textbf{Channels} & \textbf{Image Resolution} & \textbf{No of Images} & \textbf{Classes} & \textbf{Urban} & \textbf{Rural} & \textbf{Off-terrain}\\
\hline
\multirow{2}{*}{\textbf{Hyko v2}~\cite{winkens2017hyko}~\textsuperscript{1}} & \multirow{2}{*}{2017} & VIS: 470-630 & 15 & 254x510 & 163 & 10 & \checkmark & \checkmark & \texttimes\\
 &  & NIR: 630-975 & 25 & 214x407 & 78 & 10 & \texttimes & \texttimes & \texttimes\\
\textbf{HSI-Drive}~\cite{gutierrez2023hsi} & 2023 & 600-975 & 25 & 209x416 & 752 & 9 & \checkmark & \texttimes & \texttimes \\
\textbf{H-City}~\cite{shen4560035urban}~\textsuperscript{2} & 2022 & 450-950 & 128 & 1422x1889 & 1,330 & 19 & \checkmark & \texttimes & \texttimes \\
\textbf{HSI Road}~\cite{lu2020hsi}~\textsuperscript{3} & 2020 & 680-960 & 25 & 384x192 & 3,799 & 2 & \texttimes & \checkmark & \checkmark \\

\hline
\end{tabular*}
\begin{flushleft}
\textsuperscript{*} Current datasets for urban perception (HyKo-VIS, HSI-Drive, H-City) remain limited in size, class diversity, and variations in weather and locations. \\
\textsuperscript{1} NIR Camera-based subset: Captured the immediate road area. \quad \textsuperscript{2} The only dataset with co-registered RGB. \quad \textsuperscript{3} Only two classes. 
\end{flushleft}
\label{tab:HSIDatasetsDetails}
\vspace{-2ex}
\end{table*}

\begin{table*}[!h]
\centering
\caption{Class distribution across HSI-Drive and HyKo-VIS datasets: Highlighting Non-Standard Class Labels Format and Class-Wise Pixel Distributions, leading to underrepresentation and highly imbalanced data for model training.}
\begin{tabular}{lcccc!{\vrule width 0.6pt}lcccc}
\toprule
\multicolumn{5}{c}{\textbf{HSI-Drive}} & \multicolumn{5}{c}{\textbf{HyKo-VIS}} \\
\hline
\textbf{Classes} & \textbf{Instances} & \textbf{\%} & \textbf{Pixels\textsuperscript{a}} & \textbf{\%} &
\textbf{Classes} & \textbf{Instances} & \textbf{\%} & \textbf{Pixels\textsuperscript{a}} & \textbf{\%} \\
\hline
Road          & 748  & 0.15 & 26.58 & 0.61  & Road               & 235  & 0.18 & 11.7  & 0.48 \\
Road Marks    & 736  & 0.15 & 1.32  & 0.03  & Lane Markers       & 134  & 0.10 & 0.32  & 0.01 \\
Vegetation    & 659  & 0.13 & 9.30  & 0.21  & Vegetation         & 157  & 0.12 & 2.53  & 0.10 \\
P. Metal\textsuperscript{b}  & 688  & 0.14 & 0.94  & 0.02  & Vehicles\textsuperscript{f}   & 97   & 0.08 & 0.58  & 0.02 \\
Sky           & 530  & 0.11 & 2.51  & 0.06  & Sky                & 158  & 0.12 & 2.73  & 0.11 \\
Conc.\textsuperscript{c}      & 481  & 0.10 & 2.29  & 0.05  & Building, Walls    & 122  & 0.09 & 1.63  & 0.07 \\
Pedestrian    & 195  & 0.04 & 0.21  & 0.01  & Pedestrian\textsuperscript{g}   & 22   & 0.02 & 0.06  & 2e-4 \\
Water         & 2    & 4e-4 & 0.01  & 2e-4  & Sidewalk           & 107  & 0.08 & 1.18  & 0.05 \\
U. Metal\textsuperscript{d}  & 396  & 0.08 & 0.35  & 0.01  & Signs\textsuperscript{h}       & 101  & 0.08 & 0.28  & 0.01 \\
Glass\textsuperscript{e}      & 494  & 0.10 & 0.25  & 0.01  & Grass              & 159  & 0.12 & 3.52  & 0.14 \\
\bottomrule
\end{tabular}
\begin{flushleft}
\textsuperscript{a} In Millions. \quad
\textsuperscript{b} Painted Metal (vehicles, etc). \quad
\textsuperscript{c} Concrete and buildings. \quad
\textsuperscript{d} Unpainted Metals (poles, etc.). \quad
\textsuperscript{e} Including Transparent Plastic. \\
\textsuperscript{f} Car, Truck, Train, Bus, Bicycle, etc. \quad
\textsuperscript{g} Adult, Children, Cyclist, Motorcyclist, and Animal. \quad
\textsuperscript{h} Panels, Signs, and Traffic Lights.
\end{flushleft}
\label{tab:ClassImbalanceDistribution_HSI_Datasets}
\vspace{-2ex}
\end{table*}

\section{Current State of HSI in AD Scenarios}
\label{sec:Current State}
\noindent While research in RGB-based semantic segmentation has established key benchmarks through datasets such as Cityscapes~\cite{cordts2015cityscapes}, KAIST~\cite{hwang2015multispectral}, KITTI~\cite{geiger2012we}, and nuScenes~\cite{caesar2020nuscenes}, HSI-based segmentation in ADAS/AD is still a developing field. Recent advances in portable HSI technology, particularly snapshot cameras, have enabled its application in dynamic AD scenarios~\cite{gutierrez2022exploring}. The enhanced spectral information embedded in HSI offers the potential to advance object classification and material identification, which is crucial for robust ADAS/AD scene understanding.

\vspace{-2ex}
\subsection{HSI-based Datasets for AD Scenarios}
\label{sec:Current State - HSI Datasets}

\noindent Several urban-rural driving scenario based HSI datasets have been developed for ADAS/AD applications, each with distinct characteristics, as shown in Table~\ref{tab:HSIDatasetsDetails}. These datasets leverage HSI's capability, providing spectrally-rich information compared to conventional RGB images with just three channels, and cover visible (VIS) to near-infrared (NIR) wavelengths for each spatial pixel: 
\begin{itemize}
    \item The HyKo~\cite{winkens2017hyko} dataset comprises two distinct sub-datasets: HyKo-NIR (covering the wavelength range of 630-975nm) and HyKo-VIS (470-630nm). While HyKo-NIR's road-focused camera view limits its applicability for comprehensive ADAS/AD perception tasks, HyKo-VIS offers 163 HSI cubes with 15 spectral bands.
    \item The HSI-Drive~\cite{gutierrez2023hsi} dataset presents a more extensive collection spanning four weather seasons and various environmental conditions, although it is constrained by coarse pixel-wise class annotations and limited spectral coverage (25 bands, 600-975nm, primarily red and NIR spectra).
    \item The H-City~\cite{shen4560035urban} dataset offers the highest spectral resolution with 128 channels and 19 finely-annotated class labels.
    \item The HSI-Road~\cite{lu2020hsi} dataset provides binary classification (Road and Others).
\end{itemize}
Despite these diverse and significant efforts in dataset collection, several challenges remain across the existing HSI datasets. The common limitations of these datasets are:
\begin{itemize}
\item{Lack of standardization in spatial dimensions and spectral bands,}
\item{Varying spectral coverage across datasets,}
\item{Varying annotation schemes and class definitions,}
\item{Significant class imbalance, as demonstrated in Table~\ref{tab:ClassImbalanceDistribution_HSI_Datasets} for HSI-Drive and HyKo-VIS.}
\end{itemize}

While these limitations represent challenges, the existing datasets provide a valuable foundation for evaluating HSI capabilities in ADAS/AD applications and highlight key areas for future dataset development.

\begin{table*}[h!]
\centering
\caption{Current State of HSI in ADAS/AD Scenarios: Summary of Methodologies used on Urban HSI Datasets~\cite{shah2025hyperspectral}.}
\begin{tabularx}{\textwidth}{@{} p{3.15cm} p{0.40cm} p{4.3cm} >{\noindent\justifying\arraybackslash}X @{}}
\toprule
\textbf{Author} & \textbf{Year} & \textbf{Dataset Used} & \textbf{Method and Techniques} \\
\midrule
Basterretxea et al.~\cite{basterretxea2021hsi} & 2021 & HSI-Drive & Per-pixel ANN followed by two-staged spatial regularization \\
Gutiérrez-Zaballa et al.~\cite{gutierrez2022exploring} & 2022 & HSI-Drive & Patches-based UNet, and Per-pixel ANN \\
Gutiérrez-Zaballa et al.~\cite{gutierrez2023hsi} & 2023 & HSI-Drive & Patches-based UNet~\cite{ronneberger2015u} \\
Ding et al.~\cite{ding2023dual} & 2023 & H-City & HSI and pseudo-RGB based spectral feature fusion and multi-scale decoder \\
Theisen et al.~\cite{theisen2024hs3} & 2024 & HyKo-VIS, HSI-Drive, and H-City & RGB-HSI benchmarking framework using DeepLabv3+~\cite{chen2017deeplab} and UNet \\
Gutiérrez-Zaballa et al. & 2024 & HSI-Drive & Parameter perturbations~\cite{gutieerrez2024designing} and activation functions~\cite{gutierrez2024evaluating} impact in DNN\\
Shah et al.~\cite{shah2024hyperspectral} & 2024 & Relabeled HyKo-VIS, HSI-Drive, and H-City to 6--8 classes, and HSI-Road & DeepLabv3+, HRNet~\cite{sun2019high}, PSPNet~\cite{zhao2017pyramid}, and UNet, along with two Attention Mechanism-based UNet Variants, i.e., CBAM~\cite{woo2018cbam} and CA~\cite{dang2021coordinate} \\
Li et al.~\cite{li2025csnrjmimbasedspectral, li2025hyperspectral} & 2025 & H-City & CSNR and JMIM-based dimensionality reduction to generate pseudo-RGB, and evaluation through UNet, Deeplabv3+, and Segformer~\cite{xie2021segformer}\\
Gutiérrez-Zaballa et al.~\cite{gutierrez2025optimization} & 2025 & HSI-Drive & HSI optimized field programmable gate array deployed DNN segmentor\\
Shah et al.~\cite{shah2026learnable} & 2026 & HyKo-VIS, HSI-Drive, and H-City & Interpretable spectral response filters as dimensionality reduction method, evaluated with UNet, PSPNet, FPN~\cite{lin2017feature}, DeepLabv3+, and SegFormer \\
\bottomrule
\end{tabularx}
\label{tab:LitAndWorkOn_HSI}
\end{table*}

\subsection{HSI-based Semantic Segmentation in ADAS/AD}
\label{sec:Current State - HSI Semantic Segmentation}

\noindent Compared to other domains such as remote sensing, the application of deep learning-based HSI analysis in ADAS/AD is still in its early stages, as summarized in Table~\ref{tab:LitAndWorkOn_HSI}. The earlier studies cited in this table primarily focused on dataset generation and initial evaluation, with limited deep learning applications. More recent research (2024)  began to emphasize comprehensive evaluation across multiple datasets and to leverage advanced deep learning models. For example, Theisen et al.~\cite{theisen2024hs3} introduced ‘HS3-Bench’, providing a standardized evaluation framework, establishing baseline performances for semantic segmentation models (SSM) by evaluating DeepLabv3+~\cite{chen2017deeplab} and UNet on HyKo-VIS, HSI-Drive, and H-City. Shah et al.~\cite{shah2024hyperspectral} provided a comprehensive evaluation of the prominent baseline SSMs, including DeepLabv3+, HRNet~\cite{sun2019high}, PSPNet~\cite{zhao2017pyramid}, and UNet, along with AM-based UNet variants such as Conv based Attention Module (CBAM)~\cite{woo2018cbam} and Coordinate Attention (CA)~\cite{dang2021coordinate}. These efforts show the growing emphasis on the use of HSI for ADAS/AD and its potential in overall scene understanding.

Recent research highlights several key challenges in leveraging HSI for ADAS/AD applications, such as: (1) The highly dynamic nature of driving scenarios makes spectral feature extraction more complex compared to relatively static remote sensing applications~\cite{shah2024hyperspectral}. (2) Existing deep learning-based methods often benefit more from RGB, suggesting that current architectures do not optimally exploit the rich spectral data~\cite{shen4560035urban}. (3) The limited availability of annotated HSI datasets for ADAS/AD scenarios constrains the development of robust spectral feature extraction methods~\cite{shah2024hyperspectral}.

\subsection{Spectral Feature Extraction Approaches}
\label{sec:Current State - Spectral}
\noindent Traditional approaches to HSI feature extraction have primarily focused on remote sensing applications, where spectral information has been shown to play a crucial role in material classification. These methods can be broadly categorized as follows:
\begin{itemize}
    \item \textbf{Classical Image Processing} methods utilize established image processing techniques to segment objects based on known characteristics. These include thresholding, clustering, watershed segmentation, morphological operations, edge detection, superpixel generation, and region-based segmentation approaches~\cite{grewal2023hyperspectral}. While effective in controlled environments, these techniques require prior domain knowledge and often lack generalizability across diverse scenarios.
    \item \textbf{Channel-wise Processing} methods analyze each spectral band independently, typically utilizing ANNs or 1D conv to capture spectral features. However, this independent processing approach does not take advantage of important inter-channel relationships and spectral-spatial correlations that are essential for comprehensive feature extraction.
    \item \textbf{Joint Spatial-Spectral Processing} methods, such as 3D conv~\cite{Maturana2015VoxNetA3} and 3D-2D CNN feature hierarchy and pyramidal residual models~\cite{wei2025joint}, which simultaneously process spatial and spectral dimensions. Although these methods allow for a more thorough integration of information, they require substantial computational power because of the high dimensionality associated with HSI data.
    \item \textbf{Attention-based Processing} leverages attention mechanisms (AM) to selectively focus on relevant spectral bands and spatial regions. AMs are implemented through various approaches: as part of 1D-2D conv architectures within Transformer frameworks~\cite{zhang2025attention}, or through multiplicative, additive, or pooling operations such as Global Average Pooling. These attention-based methods enable the network to adaptively weight and combine features for more effective representations.
\end{itemize}

While these techniques have demonstrated success in various domains, their application to dynamic ADAS/AD remains limited and requires further evaluation. As indicated in Table~\ref{tab:LitAndWorkOn_HSI}, HSI processing for ADAS/AD is still in its early research stages, and further research is needed to evaluate and advance HSI processing methods for these applications.

\begin{figure}[!h]
\centering
\includegraphics[width=0.95\columnwidth]{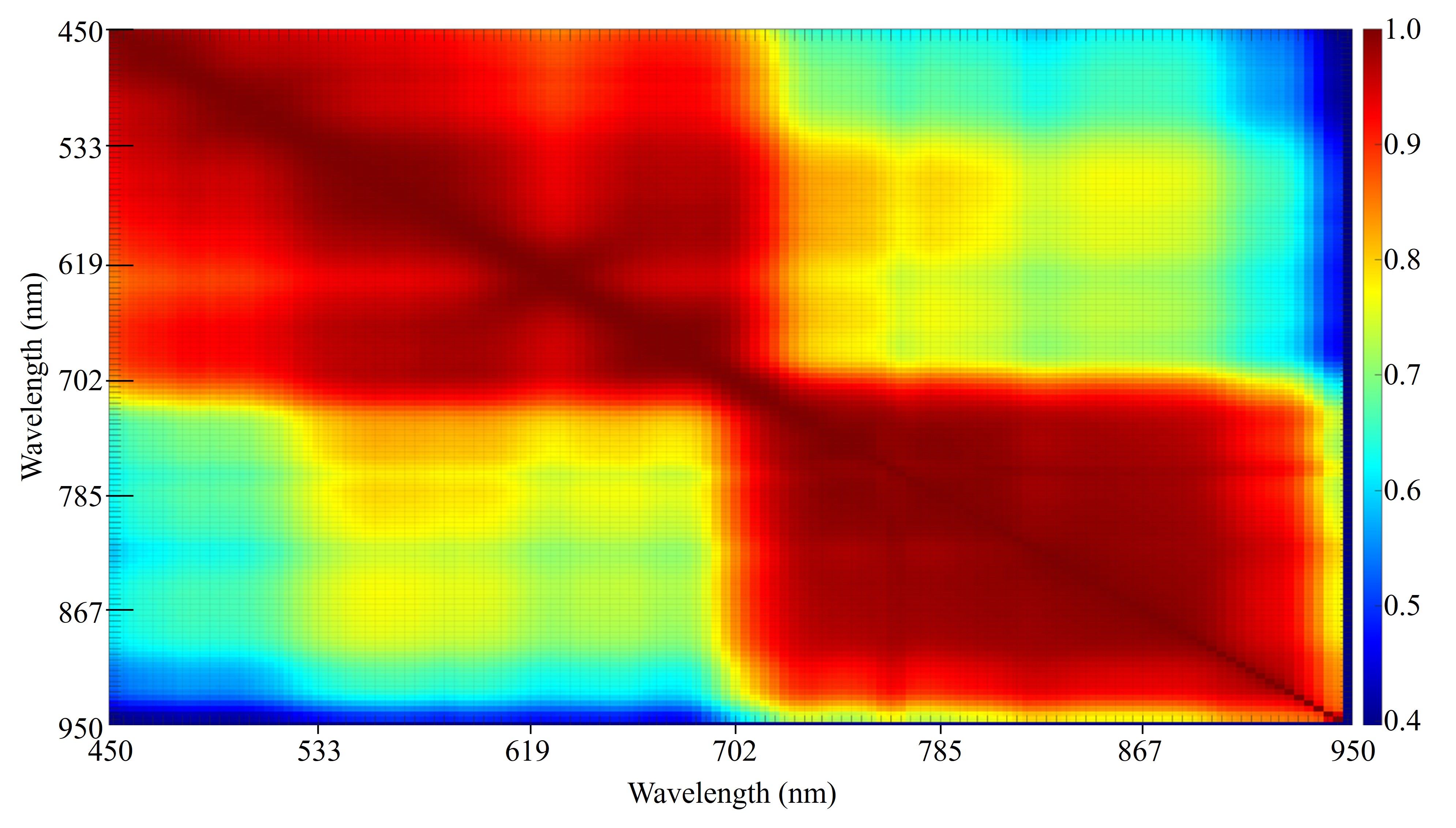}
\caption{Spectral correlation matrix across the H-City's 128 bands with strong VIS and NIR inter-band correlations.}
\label{fig:MSAM_Motivation_BandsCorrelation}
\vspace{-3ex}
\end{figure}

\begin{figure*}[!h]
\centering
\includegraphics[width=0.95\textwidth]{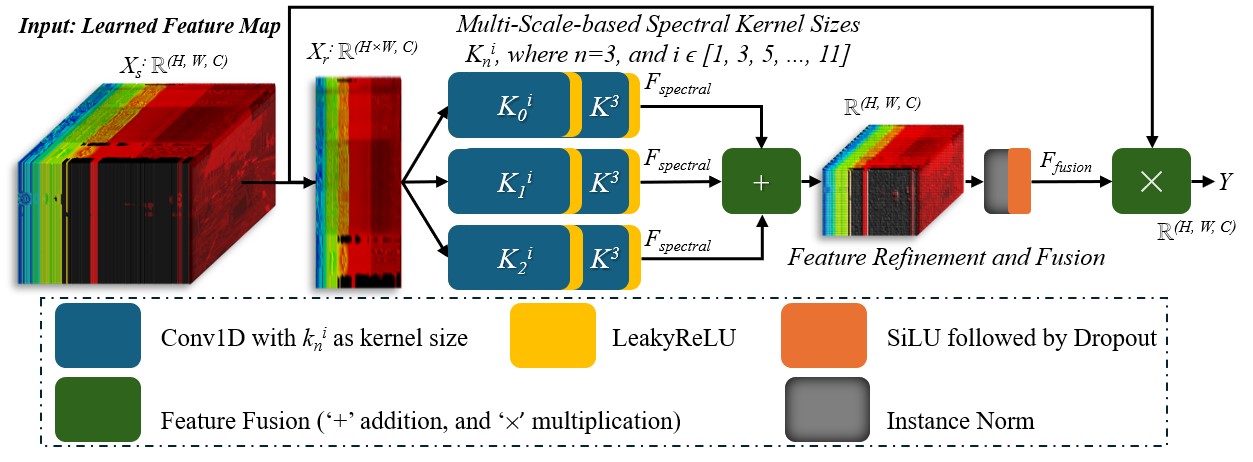}
\caption{Architecture overview of the proposed Multi-scale Spectral Attention Module (MSAM): Eq.~\eqref{Eq:1}--\eqref{Eq:4} for details.}
\label{fig:MSAM_Architecture}
\end{figure*}
\begin{figure*}[!t]
\centering
\includegraphics[width=0.95\textwidth]{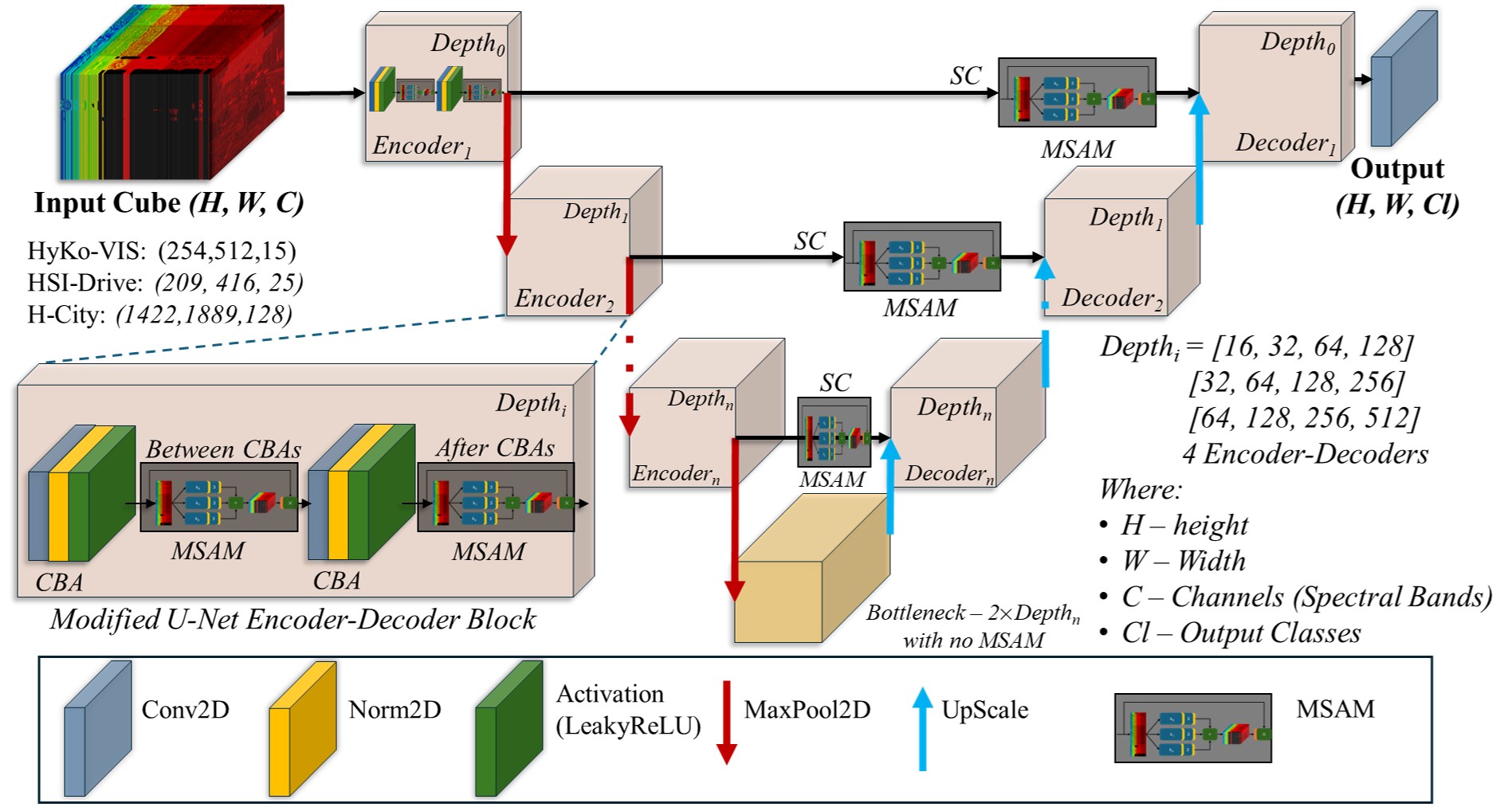}
\caption{Overview of MSAM (Fig.~\ref{fig:MSAM_Architecture}) integration in UNet architecture. MSAM is evaluated across three structural integration variants within the UNet framework: (1) positioned between the two Conv-BatchNorm-Activation (CBA) layers of the Encoder-Decoder blocks, (2) placed after the CBA layers, and (3) embedded within the skip connections (SC).}
\label{fig:UNet_MSAM_Figure}
\vspace{-4ex}
\end{figure*}

\subsection{Motivation for This Study}
\label{sec:Current State - Research Gap}
\noindent Prior HSI segmentation works in ADAS/AD either rely on per-pixel ANNs~\cite{basterretxea2021hsi} or standard patch-based UNets~\cite{gutierrez2022exploring}, which apply spatially uniform conv across all spectral bands. While Shah et al.~\cite{shah2024hyperspectral} demonstrated the benefit of attention mechanisms (CBAM~\cite{woo2018cbam}, CA~\cite{dang2021coordinate}) over baseline UNet-SC, these mechanisms operate on channel statistics (global average/max pooling) and do not explicitly model the local-to-global spectral correlation. In HSI, adjacent spectral bands contain physically correlated material responses as shown in Fig.~\ref{fig:MSAM_Motivation_BandsCorrelation}, and these relationships span different spectral scales depending on the sensor's bandwidth. A single kernel size cannot simultaneously capture narrow-band correlations and broader spectral envelopes. This motivates the investigation of multi-scale spectral filtering along the channel dimension. To address this, a Multi-scale Spectral Attention Module (MSAM) is empirically evaluated as a mechanism for capturing fine-grained spectral details and contextual spectral information at multiple scales in parallel, thereby enhancing spectral relationship modelling.

\section{Methodology}  \label{sec:Methodology}
\noindent This section outlines the architecture (Fig.~\ref{fig:MSAM_Architecture}) and integration of MSAM within the UNet framework (Fig.~\ref{fig:UNet_MSAM_Figure}).

\subsection{Multi-scale Spectral Attention Module (MSAM)} \label{sec:Methodology - MSAM}
MSAM, illustrated in Fig.~\ref{fig:MSAM_Architecture}, is designed to extract and exploit the rich spectral information of HSI while preserving spatial context. It is applied to the intermediate UNet encoder features (with $C$ learned feature channels), so its multi-scale 1D convs operate along the learned channel dimension, capturing dependencies among features derived from the spectral input. The module is based on two key principles: multi-scale spectral feature extraction and adaptive feature fusion:

\begin{equation}
Y = X_s \otimes (1 + F_{fusion}(F_{spectral}(X_r)))
\label{Eq:1}
\end{equation}

Equation~\ref{Eq:1} shows the high-level MSAM description: $\otimes$ represents element-wise multiplication; $X_s \in \mathbb{R}^{(C,H,W)}$ is the input tensor; and $X_r \in \mathbb{R}^{(H \cdot W, C)}$  is its reshaped version for channel-wise 1D conv-based feature extraction; $F_{spectral}$ and $F_{fusion}$ denote the multi-scale spectral feature extraction and the aggregation-based fusion, respectively.

\subsection{Multi-scale Spectral Feature Extraction Block} \label{sec:Methodology - Spectral Block}
Multi-scale spectral relationships are captured using three parallel 1D conv branches with different kernel sizes. Given an input containing $C$ learned feature channels, the spectral feature extraction process is defined by  Eq.~(\ref{Eq:2}):

\begin{equation}
F_{spectral} = \sigma(K^3 * \sigma(X_r * K_d^i))
\label{Eq:2}
\end{equation}

where $K$ represents the kernel, $i$ represents different kernel sizes (e.g., 3, 7, or 11 from a range of odd numbers 1 to 11), and $d$ represents the dilation rate for the 1D Conv-based on $max(1, ceil(i/2))$. $K^3$ represents a 1D kernel with a kernel size of 3, and $\sigma$ represents LeakyReLU-based activation. The proposed multi-scale approach enables capturing local learned spectral correlations (based on smaller kernels such as $i$ = 1, 3), medium-range dependencies (for $i$ = 5, 7), and broader context (for $i$ = 9, 11). The learned feature channels do not necessarily correspond one-to-one to spectral bands or retain wavelength ordering.

\subsection{Adaptive Feature Refinement Block}  \label{sec:Methodology - Adaptive Features}
The feature refinement block consists of three steps. 1) \textit{Feature Fusion}: The extracted features are aggregated through element-wise summation to capture the spectral relationships at different scales. Afterwards, Instance Normalization is used to normalize the individual channels (i.e., HxW) for better handling of spectral signature variations and to maintain stable training. 2) \textit{Addition of Non-Linearity}: To enhance feature representation, sigmoid linear unit (SiLU) activation adds non-linearity to the already fused multi-scale spectral feature maps. This step highlights the important spectral regions for further integration with spatial information by retaining fine-grained details. 3) \textit{Multiplicative Attention}: To allow the MSAM to adaptively emphasize relative spectral features, the final output is computed through multiplicative attention with the input tensor. Steps 1-2 are shown in  Eq.~(\ref{Eq:3}), and Step 3 is shown in  Eq.~(\ref{Eq:4}):

\begin{equation}
F_{fusion} = SiLU(InstanceNorm(\sum_{n=0}^{2} F_{spectral}^n)_s)
\label{Eq:3}
\end{equation}

where $n$ represents three parallel $F_{spectral}$ feature maps from Eq.~(\ref{Eq:2}), and $s$ denotes the reshape to the original input tensor shape for consistency with the network flow and integration.

\begin{equation}
Y = X_s \otimes (1 + F_{fusion})
\label{Eq:4}
\end{equation}

where $Y$ represents the refined features based on multiplicative attention of spatial feature maps with multi-scale based spectral features.

\subsection{Integration with UNet Architecture}  \label{sec:Methodology - UNet Integration}
UNet was selected as the baseline architecture because previous studies reported superior segmentation performance compared with DeepLabv3+, HRNet, and PSPNet on HSI datasets for ADAS/AD applications~\cite{shah2024hyperspectral}. To investigate the influence of module placement, MSAM was evaluated at three integration locations within UNet, as illustrated in Fig.~\ref{fig:UNet_MSAM_Figure}: (1) between consecutive CBA blocks, (2) after the CBA blocks within encoder and decoder stages, and (3) within the skip connections (SCs).

Based on the ablation studies (described in Section~\ref{sec:Experiments}--\ref{sec:Experiments - Ablative Studies}), MSAM in SC outperforms the other configurations because it enhances the flow of spectral information from the encoder to the decoder blocks of the UNet model, helping to preserve fine-grained spectral details. The integration of MSAM in UNet-SC, i.e., UNet-MSAM, maintains the UNet model backbone structure of encoder-decoder blocks, while using MSAM in SC helps with spectral information flow at multiple levels of encoder-decoder blocks.

The proposed UNet-MSAM allows the network to (1) preserve fine-grained spectral details through SC, (2) maintain computational efficiency, and (3) enable adaptive feature refinement at multi-level spectral-spatial-based fusion of features due to UNet’s use of MaxPool2D and UpSample operations at the end of each encoder and start of decoder blocks, respectively.

\begin{table}[!t]
\centering
\small
\caption{Hyperparameters used for MSAM Evaluation}
\begin{tabularx}{0.495\textwidth}{p{0.12\textwidth}ccc}
\toprule
\textbf{Datasets}              & \textbf{HyKo-VIS}          & \textbf{HSI-Drive}          & \textbf{H-City**}\\ 
\midrule
Input Size                     & 254x510x15             & 209x416x25             & 355x472x128\\ 
Dataset Size                   & 163                    & 752                    & 1,330\\ 
Batch and Acc*                 & 16, 2                   & 16, 2                   & 8, 4\\ 
Original Classes               & 10                     & 9 (less water)         & 18 (less train)\\ 
Relabeled Classes              & 7                      & 7                      & 6\\ 
\midrule
LR, AF and Epoch          & \multicolumn{3}{c}{7e-4, LeakyReLU~\cite{maas2013rectifier}, 300}\\
Loss Function                  & \multicolumn{3}{c}{Dice Loss and Class-weighted Cross Entropy}\\
Optimizer                      & \multicolumn{3}{c}{AdaBelief~\cite{zhuang2020adabelief} (beta1: 0.9 and beta2: 0.99)}\\ 
Scheduler                      & \multicolumn{3}{c}{ReduceOnPlateau (factor: 0.91, min: 5e-7)}\\ 
Hardware                       & \multicolumn{3}{c}{Core i9-13900K with Nvidia RTX 4090TI}\\ 
\bottomrule
\end{tabularx}

\begin{flushleft}
\footnotesize
* LR: Learning Rate, Acc: Accumulation Step, AF: Activation Function. \\
** 4 subsamples of original dimensions, reducing computations.
\end{flushleft}
\label{tab:Experimental_Setup}
\vspace{-2ex}
\end{table}

\begin{figure}[!t]
\centering
\subfloat{
    \includegraphics[width=0.49\textwidth]{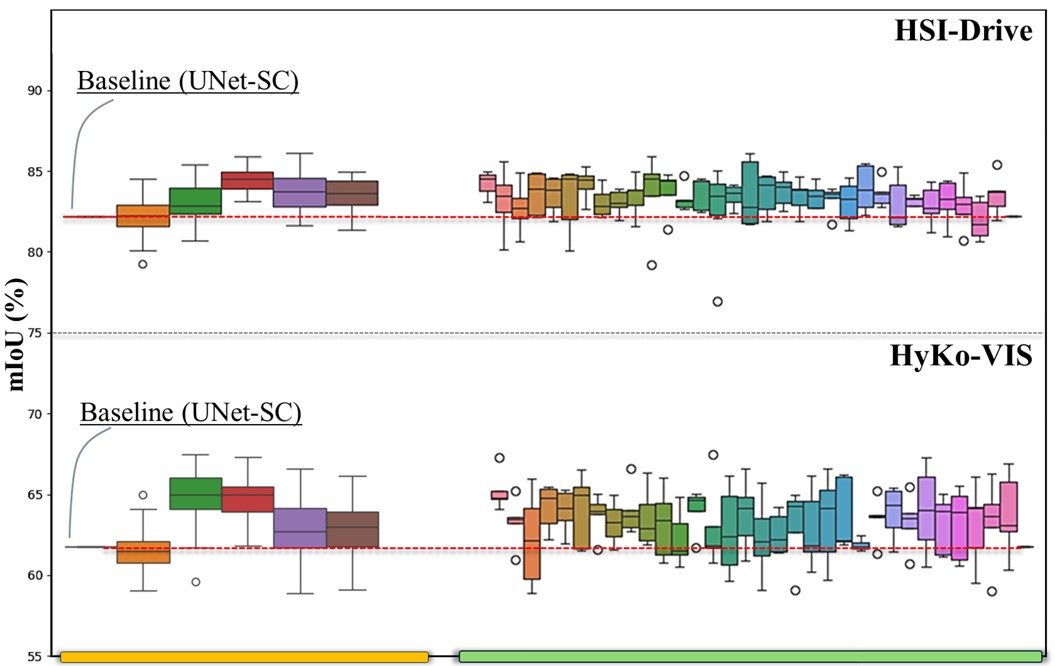}
}\\
\vspace{0.25cm}
\subfloat{
    \includegraphics[width=0.49\textwidth]{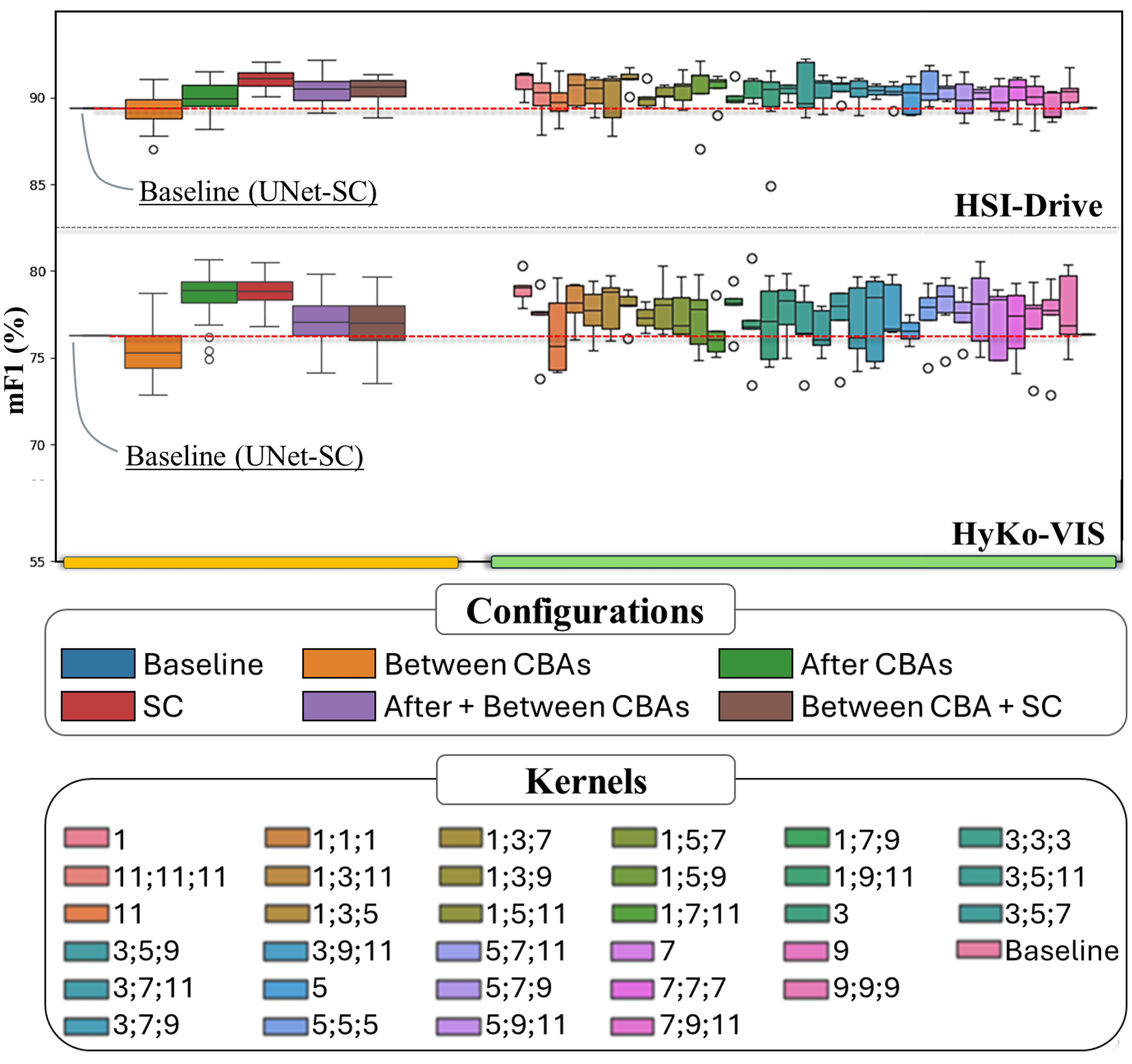}
}
\caption{mIoU (top) and mF1 (bottom) on the HSI-Drive and HyKo-VIS datasets. The yellow bars represent different MSAM integration configurations, while the green bars correspond to various kernel combinations. The red dashed line indicates the UNet-SC baseline performance.}
\label{fig:mIoU_And_mF1_performance_Ablation}
\vspace{-2.5ex}
\end{figure}

\begin{figure}[!t]
\centering
\subfloat{}
    \small
    \text{}{
    \includegraphics[width=0.49\textwidth]{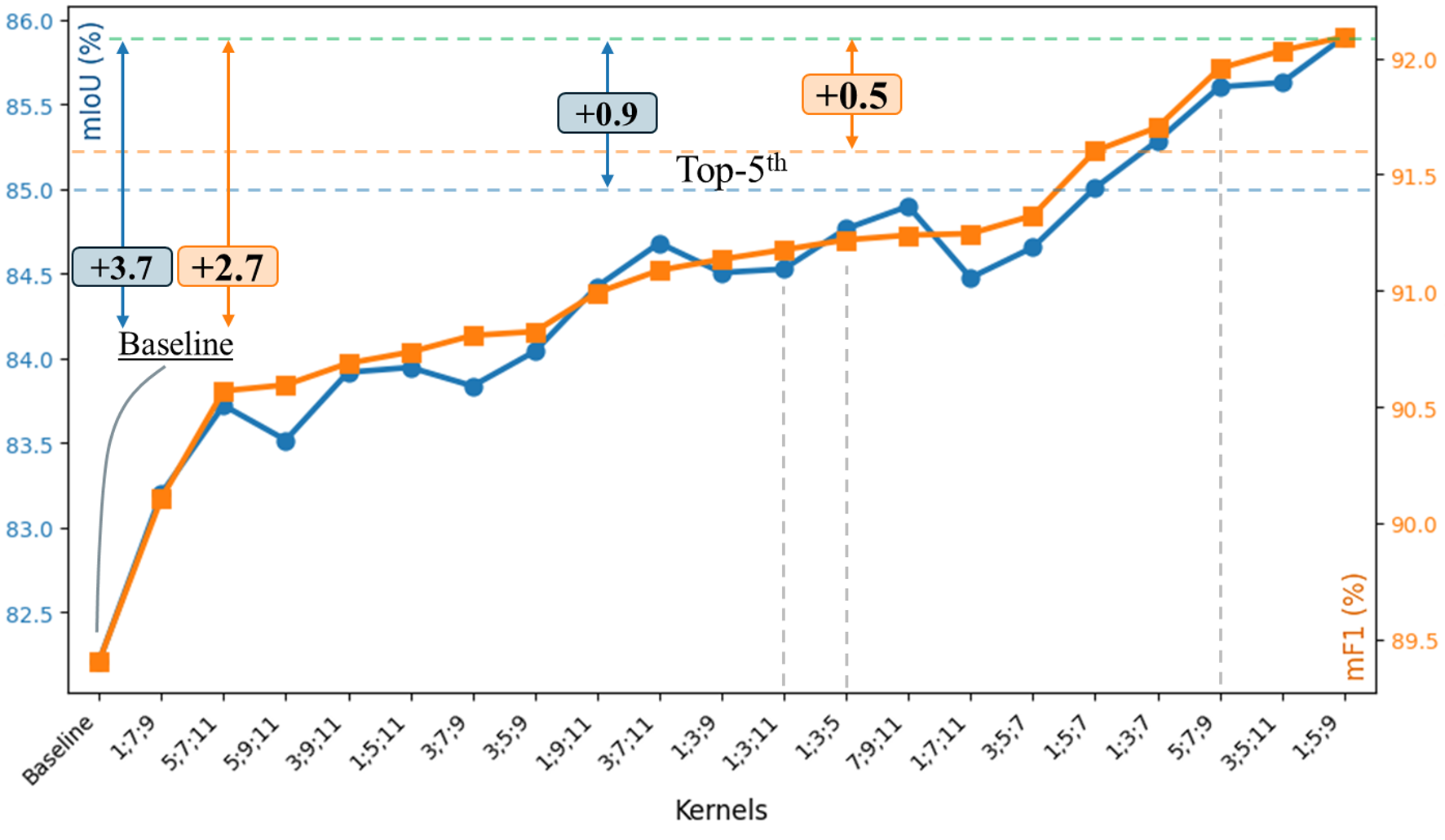}
    \label{fig:Final_HSI-Drive_kernels}
}
\vspace{-0.75cm}
\subfloat{}
    \text{}{
    \includegraphics[width=0.49\textwidth]{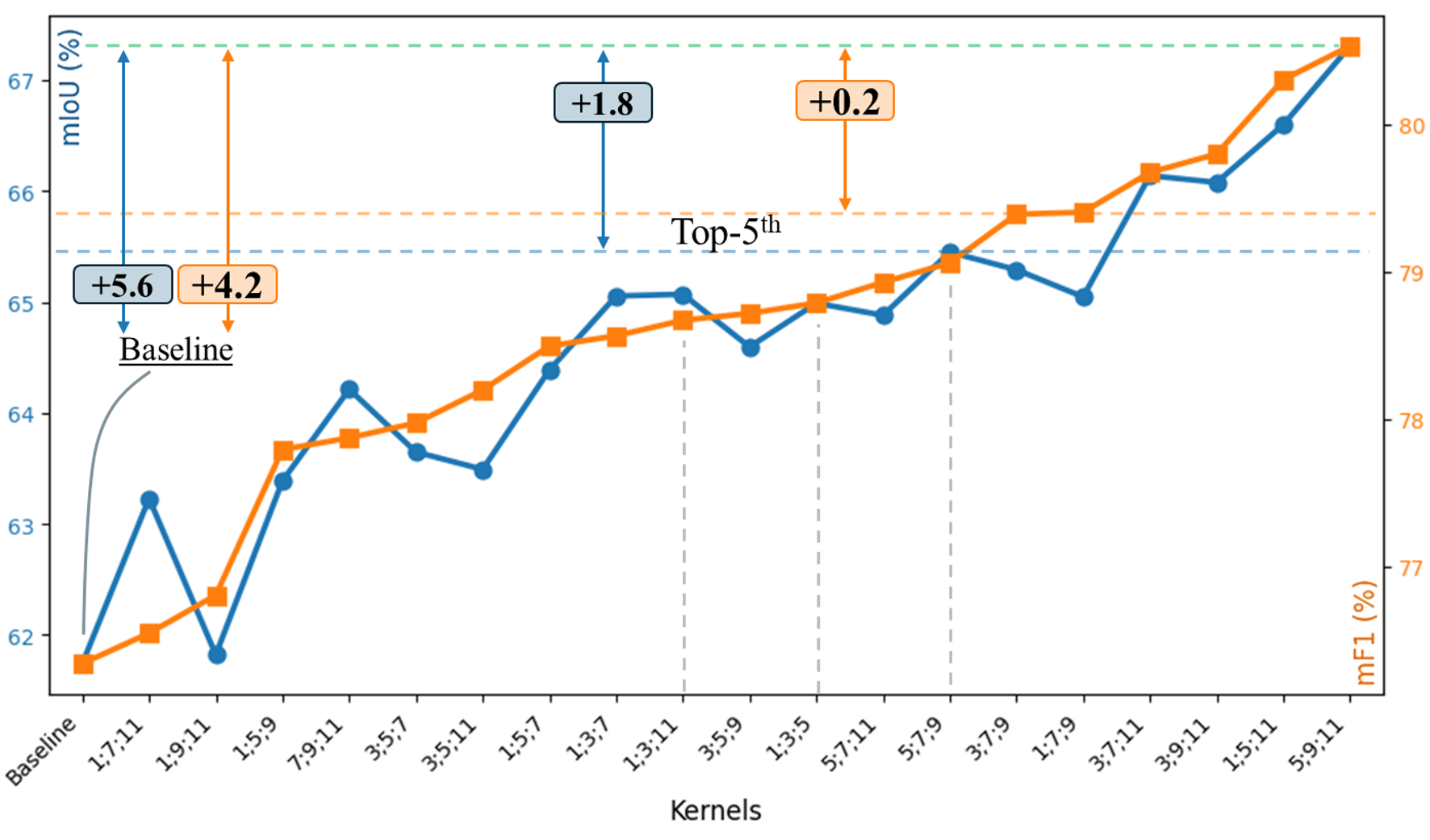}
    \label{fig:Final_HyKo-VIS_kernels}
}
\vspace{-0.55cm}
\caption{Multi-scale based kernels performance: HSI-Drive (top) and HyKo-VIS (bottom). \textit{(1;3;11)}, \textit{(1;3;5)}, and \textit{(5;7;9)} show consistent performance for both datasets.}
\label{fig:fig:Multi_Scale_Kernel_Plot}
\vspace{-2.5ex}
\end{figure}

\begin{figure}[!h]
\centering

\subfloat{}
    \text{}{
    \includegraphics[width=0.49\textwidth]{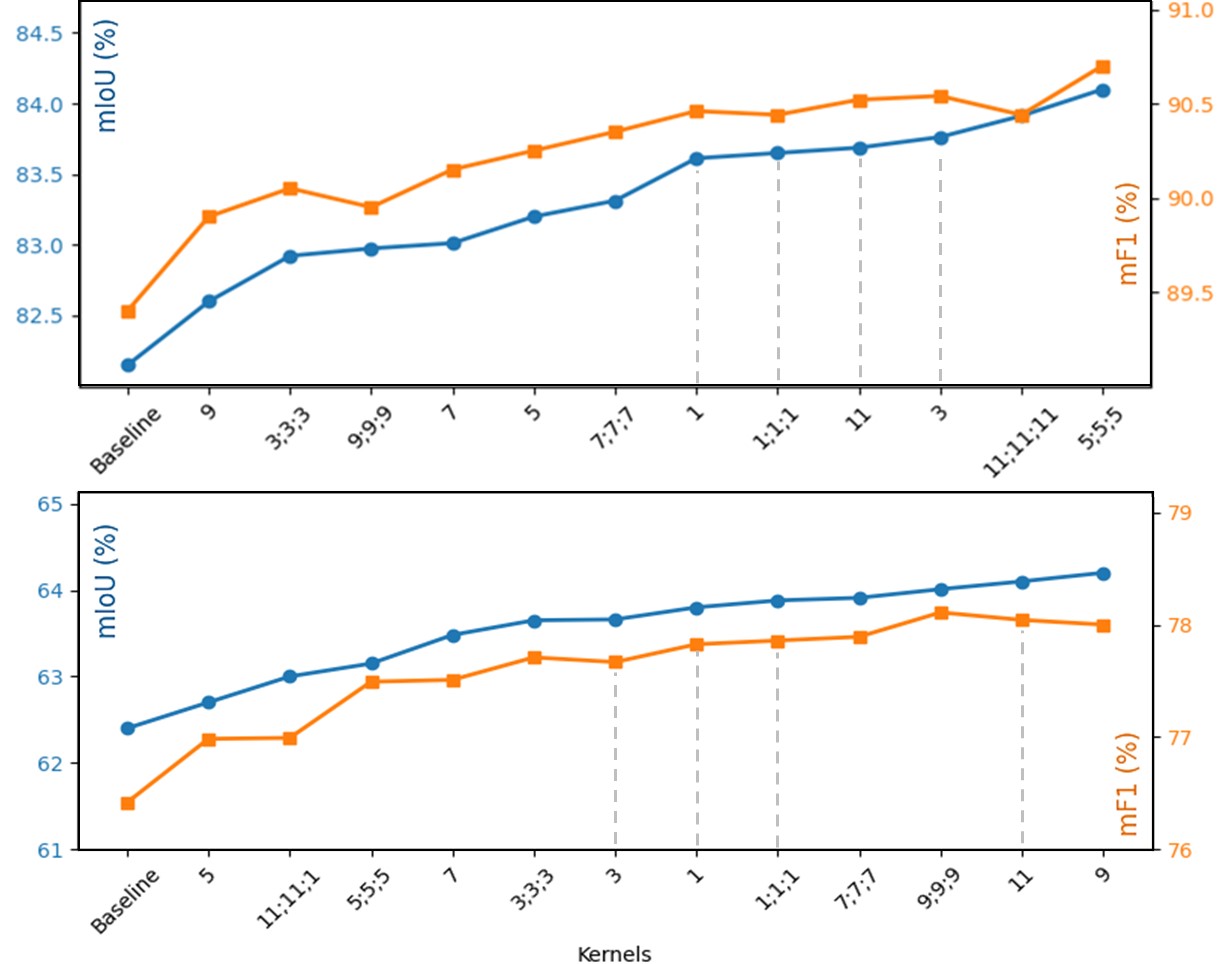}
}
\vspace{-0.55cm}
\caption{Single or same scale based kernels performance: HSI-Drive (top) and HyKo-VIS (bottom). \textit{(1)}, \textit{(11)}, and \textit{(1;1;1)} show consistent performance for both datasets.}
\label{fig:Same_Single_Kernel_Plot}
\vspace{-2.5ex}
\end{figure}

\begin{table}[!t]
\centering
\caption{Relabeled HSI Datasets$^*$: Pixelwise (in millions)}
\begin{tabular}{lrrrrrr}
\toprule
 \textbf{Class Labels} & \multicolumn{2}{c}{\textbf{HyKo-VIS}} & \multicolumn{2}{c}{\textbf{HSI-Drive}} & \multicolumn{2}{c}{\textbf{H-City**}} \\
\cmidrule(lr){2-3} \cmidrule(lr){4-5} \cmidrule(lr){6-7}
 & \textbf{Count} & \textbf{\%} & \textbf{Count} & \textbf{\%} & \textbf{Count} & \textbf{\%} \\
\midrule
Road          & 11.73  & 47.74 & 26.58 & 61.10 & 154.49 & 33.87 \\
Vegetation    & 6.06   & 24.67 & 9.3   & 21.38 & 2.00   & 0.44  \\
Sky           & 2.73   & 11.11 & 2.51  & 5.77  & 0.901  & 0.20  \\
Metal         & 0.861  & 3.50  & 1.29  & 2.97  & 153.16 & 33.58 \\
Infrastructure & 2.81   & 11.44 & 2.29  & 5.26  & 143.86 & 31.54 \\
Pedestrians        & 0.058  & 0.24  & 0.21  & 0.48  & 1.7    & 0.37  \\
Road Marking  & 0.32   & 1.30  & 1.32  & 3.03  & -      & -     \\

\bottomrule
\end{tabular}
\begin{flushleft}
\footnotesize
$^*$~Unlabeled pixels not included.~\quad $^{**}$~No Road markings in dataset.
\end{flushleft}
\label{tab:Relabeled_class_distribution}
\vspace{-2ex}
\end{table}

\section{Experimentation and Results}
\label{sec:Experiments}
\subsection{Datasets and Experimental Setup}
\label{sec:Experiments - Dataset & Setup}

The proposed multi-scale feature extraction framework is evaluated on three publicly available, multi-class annotated HSI datasets for urban driving scenarios: HyKo-VIS, HSI-Drive, and H-City, as shown in Table~\ref{tab:HSIDatasetsDetails} and discussed in Section~\ref{sec:Current State}--\ref{sec:Current State - HSI Datasets}. All models are trained using identical hyperparameters listed in Table~\ref{tab:Experimental_Setup}, with dataset splits following the original dataset releases to ensure consistency with prior work. In addition, minimal preprocessing operations were applied: (1) min-max normalization applied by the model to input tensors; (2) spatial subsampling of H-City hypercubes due to their large dimensionality, consistent with work by Shah et al.~\cite{shah2024hyperspectral}; and (3) evaluation using both relabeled annotations for ablation studies and original dataset labels for comparative analysis. The relabeled annotations are shown in Table~\ref{tab:Relabeled_class_distribution}, which consists of six common classes (Road, Vegetation, Sky, Metal, Infrastructure, and Pedestrians) and one additional dataset-specific class, i.e., Road Marking for HyKo-VIS and HSI-Drive only. This brings the number of classes to six for H-City and seven each for HyKo-VIS and HSI-Drive datasets, respectively.

The hyperparameters used in model training are listed in Table~\ref{tab:Experimental_Setup}. Given the highly imbalanced class distribution, as shown in Table~\ref{tab:Relabeled_class_distribution}, dice loss and class-weighted Cross-Entropy were used as loss functions~\cite{yeung2022unified}. The model performance was evaluated using class-wise mean-based metrics including mean Intersection over Union (mIoU) and mean F1 Score (mF1).

\subsection{Stage~1 -- MSAM Integration into UNet}
\label{sec:Experiments - Ablative Studies}
The experimental analysis addresses two critical aspects: (1) determining optimal MSAM integration locations within the UNet architecture, and (2) investigating effective kernel combinations for multi-scale spectral feature extraction. Since MSAM employs three kernels to extract fine-grained details and capture local-to-global spectral context, ablation studies comprehensively evaluated kernel size combinations ranging from 1 to 11.

Ablations were conducted using UNet backbones with four encoder-decoder blocks (depths: 32, 64, 128, 256) across five integration configurations: (1) between CBAs, (2) after CBAs, (3) in SC, (4) between+after CBAs, and (5) between CBA+SC, as discussed in Section~\ref{sec:Methodology}--\ref{sec:Methodology - UNet Integration} and illustrated in Fig.~\ref{fig:UNet_MSAM_Figure}. Each configuration was tested with 32 different kernel combinations, totaling 161 configurations per dataset. Due to computational constraints, ablation studies were conducted on HSI-Drive and HyKo-VIS datasets, with H-City reserved for final evaluation.

\textbf{Integration Location Analysis}: Fig.~\ref{fig:mIoU_And_mF1_performance_Ablation} demonstrates that MSAM integration within SC (UNet-MSAM) yields the most stable performance without outliers, consistently outperforming the baseline UNet-SC (red dashed line). This configuration maintains optimal spectral-spatial feature fusion without disrupting the spatial processing pipeline. The second-best configuration, 'After CBAs', showed higher variance due to its interaction with MaxPool2D layers in encoder blocks.  The results confirm that proper fusion of spectral and spatial features is crucial for optimal HSI segmentation.

\textbf{Kernel Combination Analysis}: All UNet-MSAM configurations outperformed baseline UNet-SC, as shown in Fig.~\ref{fig:mIoU_And_mF1_performance_Ablation}–\ref{fig:Same_Single_Kernel_Plot}:
\begin{itemize}
    \item{\textbf{Multi-Scale Kernels}: Top-performing configurations achieved improvements of 3.7\% and 5.6\% in mIoU, and 2.7\% and 4.2\% in mF1 for HSI-Drive and HyKo-VIS, respectively. The top-5 performing kernels indicate that the selection of the appropriate kernels is dataset-specific for optimal feature extraction. For HSI-Drive, the top-performing combination of \textit{(1;5;9)} achieves an improvement of 0.62\% in mIoU and 0.41\% in mF1 over the top-5th combination \textit{(1;5;7)}. Whereas for HyKo-VIS, \textit{(5;9;11)} provided 1.32\% in mIoU and 0.91\% in mF1 improvements over the top-5th combination \textit{(1;7;9)}.}
    \item{\textbf{Single/Same Scale Kernels}: Uniform kernel combinations (\textit{(1)}, \textit{(1;1;1)}, \textit{(3)}, \textit{(3;3;3)}, etc.) outperformed the baseline UNet-SC but remained well below the top-5 multi-scale combinations for the respective datasets. This indicates that the multi-scale spectral approach is more effective in preserving spectral features for HSI segmentation.}
\end{itemize}

\begin{table*}
\centering
\caption{UNet-MSAM performance across UNet depths: underlined-bold indicate the overall best mIoU and mF1 results.}
\resizebox{0.98\textwidth}{!}{%
\begin{tabular}{llcc|cc|cc}
\toprule
\multirow{1}{*}{\textbf{Dataset}} & \textbf{Configuration} & \multicolumn{2}{c|}{\textbf{UNet$_{16}$: (16, 32, 64, 128)}} & \multicolumn{2}{c|}{\textbf{UNet$_{32}$: (32, 64, 128, 256)}} & \multicolumn{2}{c}{\textbf{UNet$_{64}$: (64, 128, 256, 512)}} \\
& & \textbf{mIoU} & \textbf{mF1} & \textbf{mIoU} & \textbf{mF1} & \textbf{mIoU} & \textbf{mF1} \\
\hline
\multirow{1}{*}{HyKo-VIS} & UNet-SC & 54.33 & 69.79 & 61.78 & 70.34 & 62.53 & 76.81 \\
& \multirow{9}{*}{\rotatebox{90}{UNet-MSAM}} \hspace{0.1em} (1;3;11) & 61.58 & 76.85 & 65.42 & 79.47 & 64.79 & 79.02 \\
& \hspace{1.5em}(1;3;5) & 59.11 & 74.13 & 65.25 & 79.43 & 66.68 & 80.21 \\
& \hspace{1.5em}(1;5;11) & 60.28 & 75.46 & 65.30 & 78.88 & 68.17 & 80.67 \\
& \hspace{1.5em}(1;5;9) & 61.08 & 75.94 & 66.61 & 80.16 & 67.17 & 80.35 \\
& \hspace{1.5em}(3;5;11) & 57.37 & 72.66 & 66.67 & 79.86 & 67.36 & 80.51 \\
& \hspace{1.5em}(3;7;11) & 61.17 & 76.00 & 65.11 & 78.94 & \textbf{\underline{68.44}} & \textbf{\underline{81.50}} \\
& \hspace{1.5em}(5;7;9) & 60.57 & 75.69 & 64.88 & 78.85 & 66.24 & 79.99 \\
& \hspace{1.5em}(5;9;11) & 62.86 & 77.32 & 64.90 & 78.77 & 66.23 & 79.63 \\
\cline{2-8}
\noalign{\vskip 1pt}
& \text{UNet-MSAM} $(\mu\pm\sigma)$$^{*}$  & 
60.50 $\pm$ 1.55 &
75.51 $\pm$ 1.39 &
65.52 $\pm$ 0.67 &
79.30 $\pm$ 0.49 &
66.89 $\pm$ 1.10 &
80.24 $\pm$ 0.69 \\
& Avg Difference Increase (pp$^{**}$) & 6.17 & 5.72 & 3.74 & 8.96 & 4.36 & 3.43 \\
\hline
\multirow{1}{*}{HSI-Drive} & UNet-SC & 79.32 & 87.42 & 84.07 & 88.83 & 84.29 & 90.02 \\
& \multirow{9}{*}{\rotatebox{90}{UNet-MSAM}} \hspace{0.1em} (1;3;11) & 81.41 & 88.72 & 84.18 & 90.80 & 84.55 & 90.98 \\
& \hspace{1.5em}(1;3;5) & 81.63 & 89.32 & 84.50 & 91.22 & 86.23 & 92.32 \\
& \hspace{1.5em}(1;5;11) & 80.86 & 88.69 & 85.04 & 91.36 & \textbf{\underline{86.72}} & \textbf{\underline{92.45}} \\
& \hspace{1.5em}(1;5;9) & 80.47 & 88.30 & 83.70 & 90.54 & 84.91 & 90.49 \\
& \hspace{1.5em}(3;5;11) & 80.71 & 88.60 & 83.69 & 90.51 & 85.45 & 91.83 \\
& \hspace{1.5em}(3;7;11) & 81.72 & 89.06 & 83.47 & 90.21 & 86.40 & 92.18 \\
& \hspace{1.5em}(5;7;9) & 80.98 & 88.72 & 85.34 & 91.64 & 84.89 & 91.41 \\
& \hspace{1.5em}(5;9;11) & 81.39 & 89.01 & 84.89 & 91.44 & 85.80 & 91.68 \\
\cline{2-8}
\noalign{\vskip 1pt}
& \text{UNet-MSAM} $(\mu\pm\sigma)$  & 
81.15 $\pm$ 0.43 &
88.80 $\pm$ 0.30 &
84.35 $\pm$ 0.65 &
90.97 $\pm$ 0.49 &
85.62 $\pm$ 0.74 &
91.67 $\pm$ 0.64 \\
& Avg Difference Increase (pp) & 1.83 & 1.38 & 0.28 & 2.14 & 1.33 & 1.65 \\
\hline
\multirow{1}{*}{H-City} & UNet-SC & 78.75 & 86.58 & 86.08 & 90.54 & 86.44 & 91.88 \\
& \multirow{9}{*}{\rotatebox{90}{UNet-MSAM}} \hspace{0.1em} (1;3;11) & 82.11 & 88.94 & 86.01 & 91.06 & 87.51 & 92.73 \\
& \hspace{1.5em}(1;3;5) & 80.45 & 87.78 & 86.09 & 89.21 & 87.24 & 92.30 \\
& \hspace{1.5em}(1;5;11) & 78.21 & 86.47 & 87.43 & 92.19 & 87.34 & 92.47 \\
& \hspace{1.5em}(1;5;9) & 80.84 & 87.94 & 87.43 & 92.98 & 86.44 & 91.33 \\
& \hspace{1.5em}(3;5;11) & 80.57 & 88.10 & 87.30 & 91.64 & 86.90 & 92.51 \\
& \hspace{1.5em}(3;7;11) & 79.18 & 87.34 & 87.63 & \textbf{\underline{93.06}} & \textbf{\underline{87.80}} & 92.77 \\
& \hspace{1.5em}(5;7;9) & 79.50 & 87.57 & 86.95 & 90.32 & 87.03 & 92.23 \\
& \hspace{1.5em}(5;9;11) & 82.24 & 88.92 & 86.42 & 91.18 & 87.24 & 92.40 \\
\cline{2-8}
\noalign{\vskip 1pt}
& \text{UNet-MSAM} $(\mu\pm\sigma)$  & 
80.39 $\pm$ 1.31 &
87.88 $\pm$ 0.76 &
86.91 $\pm$ 0.61 &
91.46 $\pm$ 1.23 &
87.19 $\pm$ 0.38 &
92.34 $\pm$ 0.42 \\
& Avg Difference Increase (pp) & 1.64 & 1.30 & 0.83 & 0.92 & 0.75 & 0.46 \\
\hline
\noalign{\vskip 1pt}
\multicolumn{2}{c}{Overall Avg Difference Increase over UNet-SC (pp)} & \textbf{3.21} & \textbf{2.80} & \textbf{1.62} & \textbf{4.00} & \textbf{2.14} & \textbf{1.85} \\
\bottomrule
\end{tabular}
}
\vspace{-1.25ex}
\begin{flushleft}
\footnotesize
* $\mu\pm\sigma$ (mean$\pm$standard deviation) shows variability across the eight kernel configurations, not from independent runs. \quad $^{**}$pp: percentage points.
\end{flushleft}
\label{tab:QuantitativeResults}
\vspace{-2ex}
\end{table*}
\textbf{Key Findings}: Stage-1 ablations provide the following insights:
\begin{itemize}
    \item{Multi-scale kernels in UNet-MSAM consistently outperform the baseline UNet-SC model, in both mIoU and mF1 across datasets.}
    \item{Optimal kernel combination is dataset-specific, highlighting the unique spectral features in the datasets and the importance of spectral features in HSI segmentation}
    \item{Both MSAM integration location and kernel selection significantly impact performance, with SC integration providing optimal stability.}
    \item{These findings establish a foundation for future research in adaptive kernel selection strategies for HSI segmentation applications.}
\end{itemize}

\begin{table}[ht]
\centering
\caption{Quantitative and Computational performance comparison across established AM, averaged over three independent training runs: Bold and underlined values show the top and second-best performance, respectively.}
\resizebox{0.5\textwidth}{!}{
\small
\begin{tabular}{c|l|rrrrr}
\hline
\multirow{2}{*}{\textbf{\rotatebox{90}{Data}}} & \multirow{2}{*}{\shortstack{\textbf{  Backbone} \\ \textbf{  UNet$_{32}$}}} & \multicolumn{5}{c}{\textbf{Original Dataset Classes}} \\
\cline{3-7}
& & \textbf{param.\textsuperscript{a}} & \textbf{CPU\textsuperscript{b}} & \textbf{GPU\textsuperscript{b}} & \textbf{mIoU} & \textbf{mF1} \\
\hline
\multirow{6}{*}{\rotatebox{90}{HyKo-VIS}} 
& SE~\cite{hu2018squeeze}      & 7.778 & 70.43 & 4.07 & 67.27 & 79.68 \\
& CBAM~\cite{woo2018cbam}      & 7.779 & 79.55 & 45.54 & \textbf{67.39} & 79.66 \\
& CA~\cite{dang2021coordinate} & 7.779 & 73.95 & 7.04 & 66.71 & 79.50 \\
& MSAM$_{(3;7;11)}$            & 7.768 & 231.91 & 7.30  & 67.03 & \underline{80.09} \\
& MSAM$_{(1;5;11)}$            & 7.768 & 253.28 & 7.45 & \underline{67.31} & \textbf{80.11}
 \\
\hline
\multirow{6}{*}{\rotatebox{90}{HSI-Drive} }
& SE       & 7.781 & 72.51 & 3.16 & 80.17 & 88.55 \\
& CBAM     & 7.781 & 69.41 & 21.48 & 80.02 & 88.45 \\
& CA       & 7.782 & 66.70 & 5.01 & \textbf{81.49} & \textbf{90.03} \\
& MSAM$_{(3;7;11)}$ & 7.771 & 159.01 & 5.68 & \underline{80.70} & \underline{88.81} \\
& MSAM$_{(1;5;11)}$ & 7.771 & 190.95 & 5.70 & 80.07 & 88.46 \\
\hline
\multirow{6}{*}{\rotatebox{90}{H-City}}
& SE       & 7.811 & 116.87 & 5.42 & 63.78 & 74.78 \\
& CBAM     & 7.811 & 109.43 & 43.92 & 64.06 & 75.29 \\
& CA       & 7.812 & 104.24 & 9.28 & \underline{64.88} & \underline{75.39} \\
& MSAM$_{(3;7;11)}$ & 7.801 & 301.53 & 9.55 & 63.85 & 74.63 \\
& MSAM$_{(1;5;11)}$ & 7.801 & 327.50 & 9.56 & \textbf{65.05} & \textbf{75.45}  \\
\hline
\end{tabular}
}
\vspace{-2.5ex}
\begin{flushleft}
\footnotesize
\textsuperscript{a} In Millions.\quad\textsuperscript{b} Average over 100 hypercubes, in milliseconds.
\end{flushleft}
\label{tab:OtherBaselinesAndComputationalOverheadComparison}
\vspace{-3ex}
\end{table}

\vspace{-2ex}
\subsection{Stage~2 -- Evaluation Against Baseline UNet-SC}
\label{sec:Experiments - Stage2}
Based on Stage-1 findings, eight kernel combinations were evaluated in Stage-2: the top-3 performing kernels per dataset and three consistently strong configurations \textit{(5;7;9)} (also in top-3 of HSI-Drive), \textit{(1;3;11)}, and \textit{(1;3;5)}. These kernels were evaluated across three UNet backbone depths: UNet$_{16}$ (16, 32, 64, 128), UNet$_{32}$ (32, 64, 128, 256), and UNet$_{64}$ (64, 128, 256, 512), on all multi-class HSI datasets and annotations used in benchmarking work~\cite{shah2024hyperspectral}.

Table~\ref{tab:QuantitativeResults} shows that UNet-MSAM improves over UNet-SC with average gains of 2.32\% mIoU and 2.88\% mF1. HyKo-VIS achieved the largest gains (3.74--6.17\% mIoU, 3.43--8.96\% mF1), demonstrating maximum improvements while HSI-Drive and H-City remained modest. These results confirm that multi-scale spectral feature extraction enhances segmentation performance across the diverse HSI datasets.

\begin{figure*}[h!]
  \centering
  \setlength{\tabcolsep}{0.75pt} 
  \begin{tabular}{p{0.25cm}cc|cc|cc}
    & \multicolumn{2}{c|}{HyKo-VIS} & \multicolumn{2}{c|}{HSI-Drive} & \multicolumn{2}{c}{H-City}
    \\
    \raisebox{1.25\height}{\rotatebox{90}{\small RGB}} &
    \includegraphics[width=0.162\textwidth, height=0.11\textwidth]{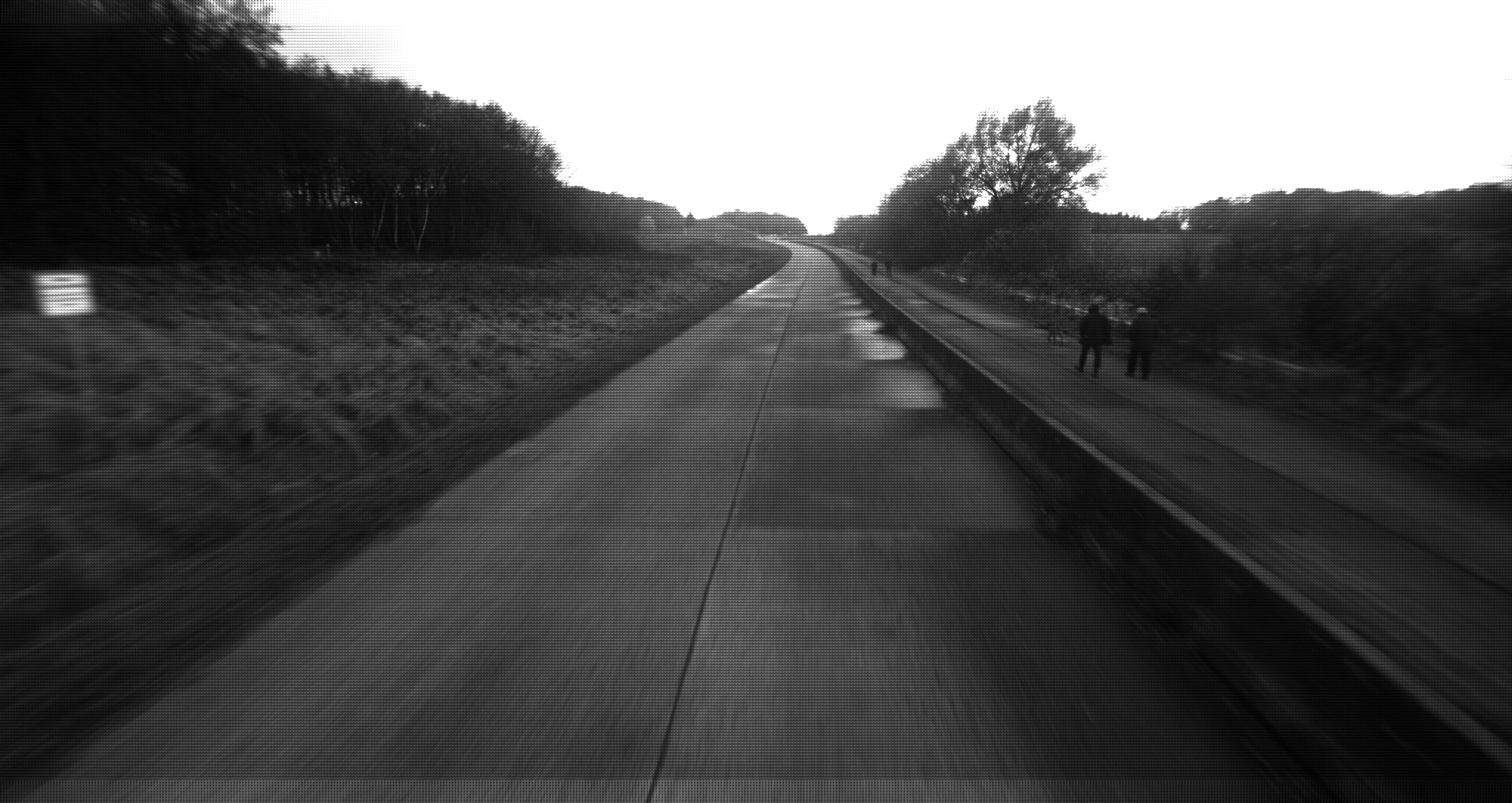} &
    \includegraphics[width=0.162\textwidth, height=0.11\textwidth]{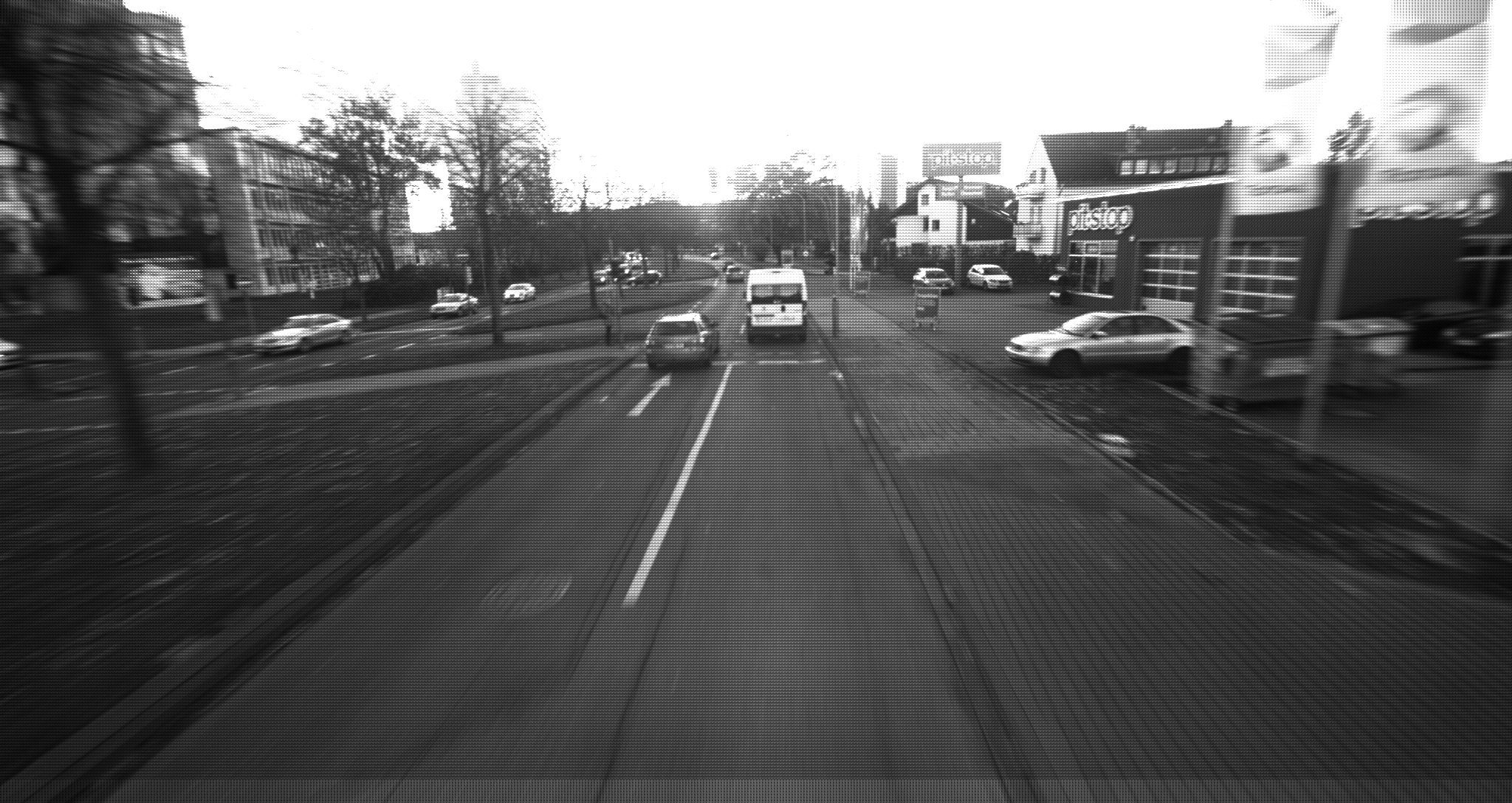} &
    \includegraphics[width=0.162\textwidth, height=0.11\textwidth]{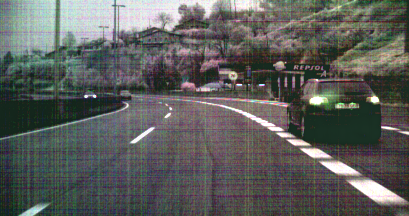} &
    \includegraphics[width=0.162\textwidth, height=0.11\textwidth]{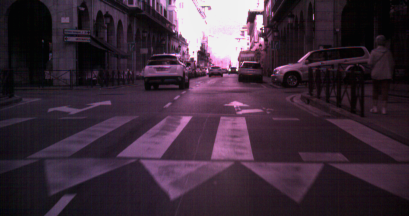} &
    \includegraphics[width=0.162\textwidth, height=0.11\textwidth]{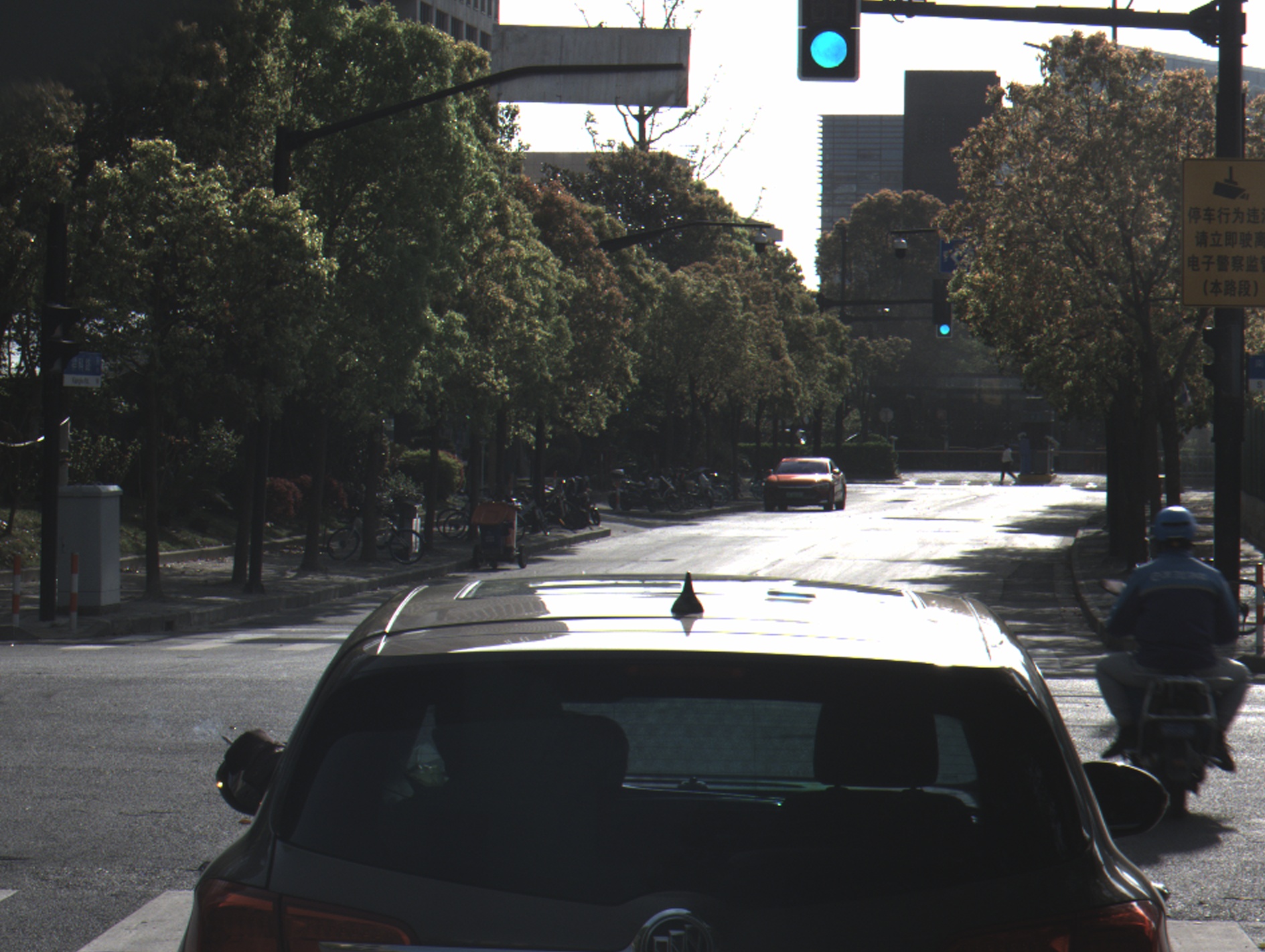} &
    \includegraphics[width=0.162\textwidth, height=0.11\textwidth]{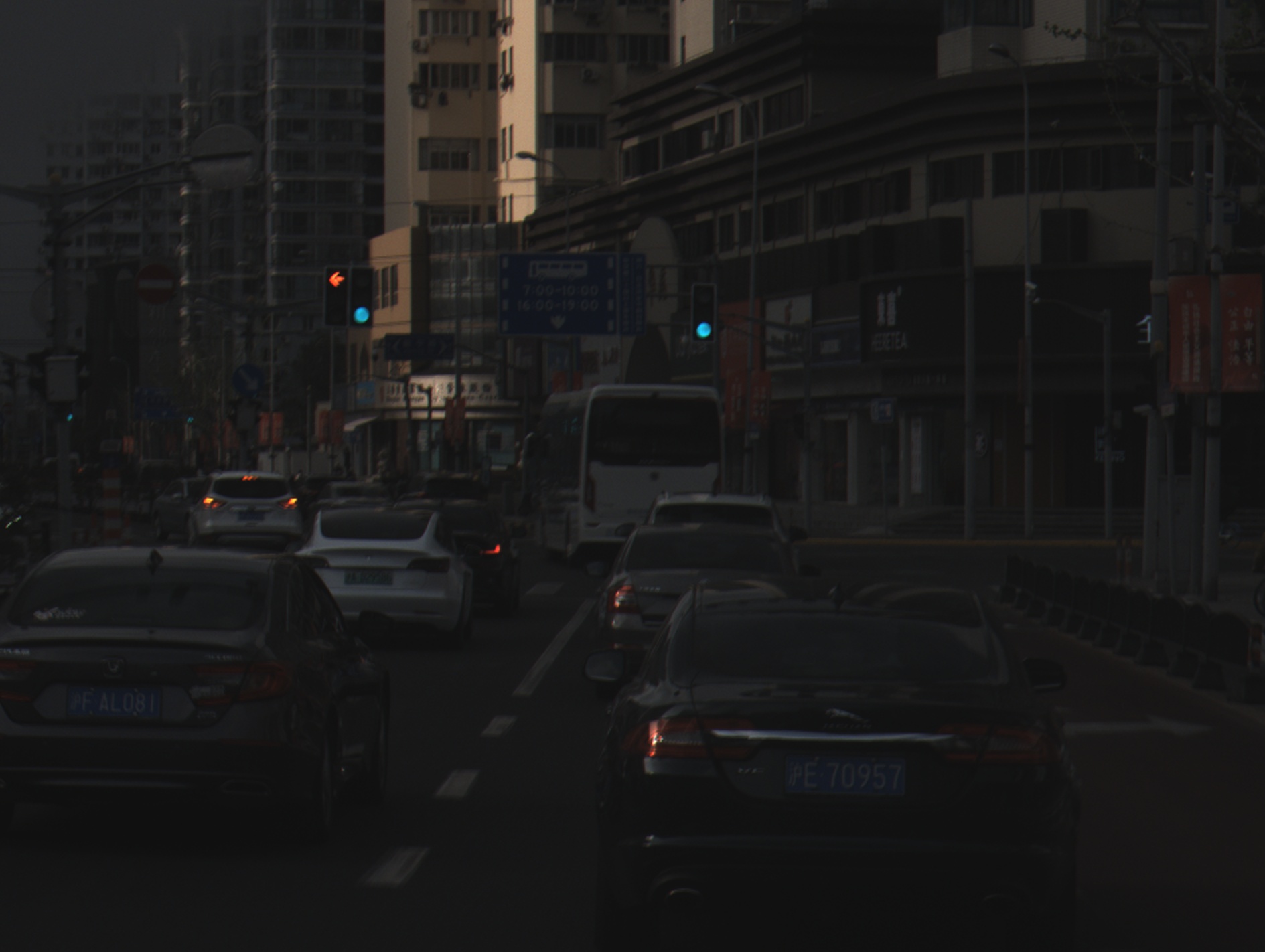} 
    \\
    \raisebox{.85\height}{\rotatebox{90}{\small Masks}} &
    \includegraphics[width=0.162\textwidth, height=0.11\textwidth]{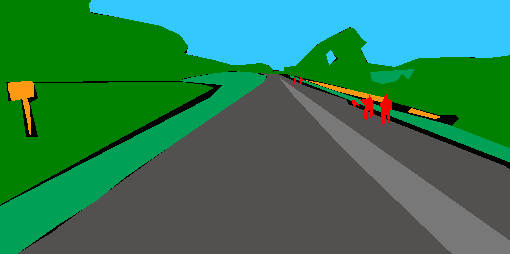} &
    \includegraphics[width=0.162\textwidth, height=0.11\textwidth]{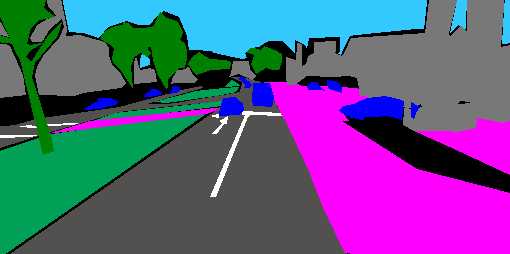} &
    \includegraphics[width=0.162\textwidth, height=0.11\textwidth]{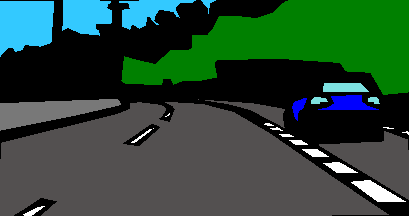} &
    \includegraphics[width=0.162\textwidth, height=0.11\textwidth]{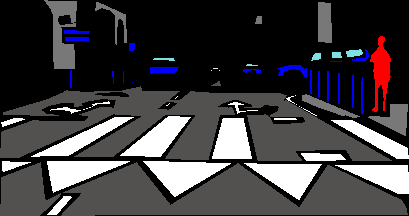} &
    \includegraphics[width=0.162\textwidth, height=0.11\textwidth]{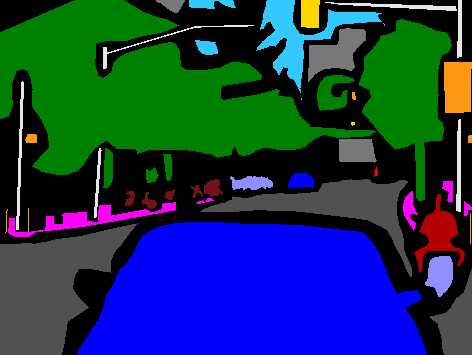} &
    \includegraphics[width=0.162\textwidth, height=0.11\textwidth]{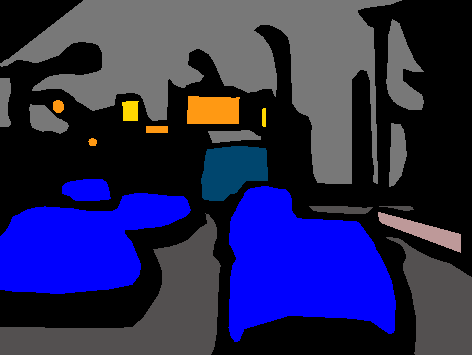}
    \\
    \raisebox{2.25\height}{\rotatebox{90}{\small SC}} &
    \includegraphics[width=0.162\textwidth, height=0.11\textwidth]{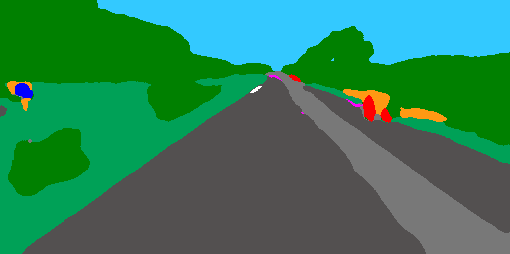} &
    \includegraphics[width=0.162\textwidth, height=0.11\textwidth]{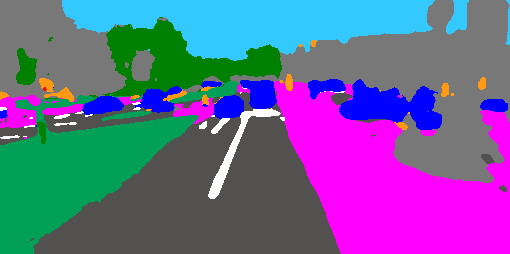} &
    \includegraphics[width=0.162\textwidth, height=0.11\textwidth]{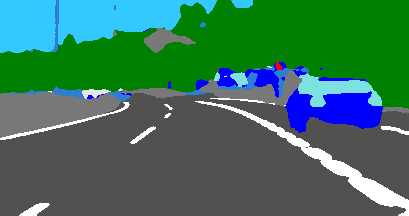} &
    \includegraphics[width=0.162\textwidth, height=0.11\textwidth]{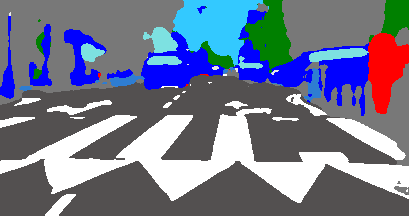} &
    \includegraphics[width=0.162\textwidth, height=0.11\textwidth]{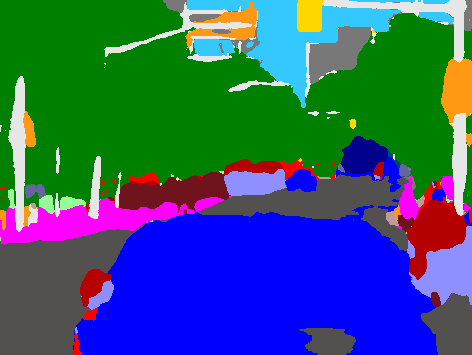} &
    \includegraphics[width=0.162\textwidth, height=0.11\textwidth]{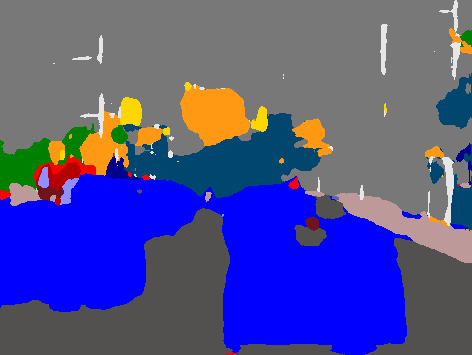}
    \\
    \raisebox{2.25\height}{\rotatebox{90}{\small SE}} &
    \includegraphics[width=0.162\textwidth, height=0.11\textwidth]{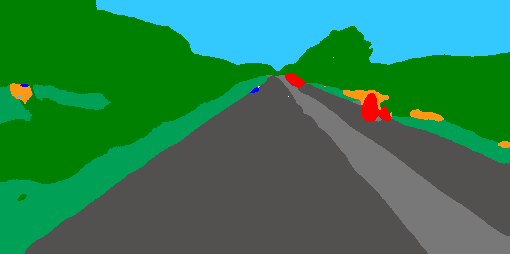} &
    \includegraphics[width=0.162\textwidth, height=0.11\textwidth]{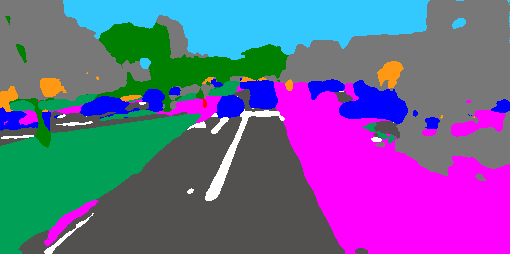} &
    \includegraphics[width=0.162\textwidth, height=0.11\textwidth]{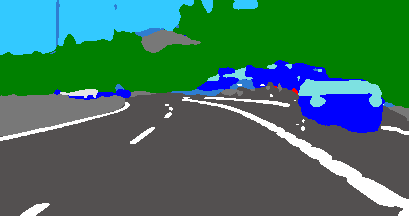} &
    \includegraphics[width=0.162\textwidth, height=0.11\textwidth]{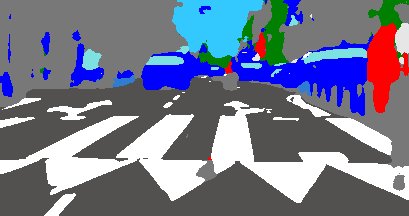} &
    \includegraphics[width=0.162\textwidth, height=0.11\textwidth]{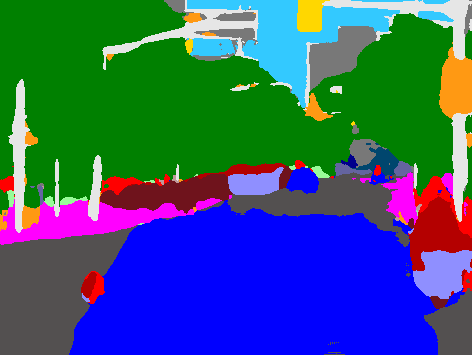} &
    \includegraphics[width=0.162\textwidth, height=0.11\textwidth]{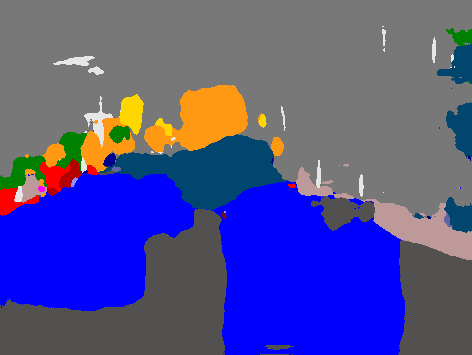}
    \\
    \raisebox{2.0\height}{\rotatebox{90}{\small CA}} &
    \includegraphics[width=0.162\textwidth, height=0.11\textwidth]{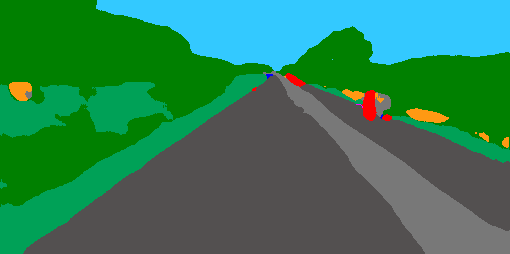} &
    \includegraphics[width=0.162\textwidth, height=0.11\textwidth]{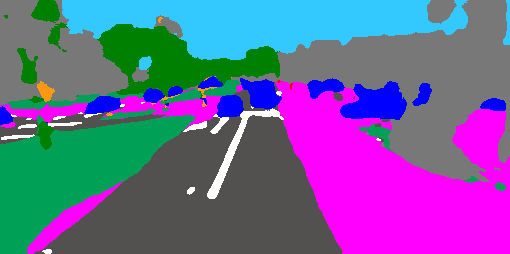} &
    \includegraphics[width=0.162\textwidth, height=0.11\textwidth]{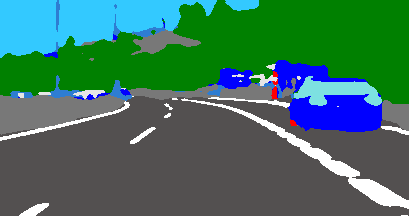} &
    \includegraphics[width=0.162\textwidth, height=0.11\textwidth]{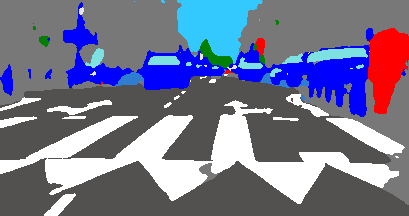} &
    \includegraphics[width=0.162\textwidth, height=0.11\textwidth]{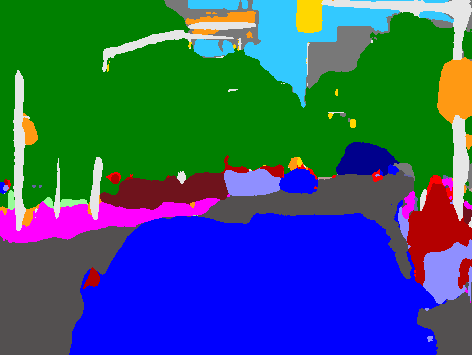} &
    \includegraphics[width=0.162\textwidth, height=0.11\textwidth]{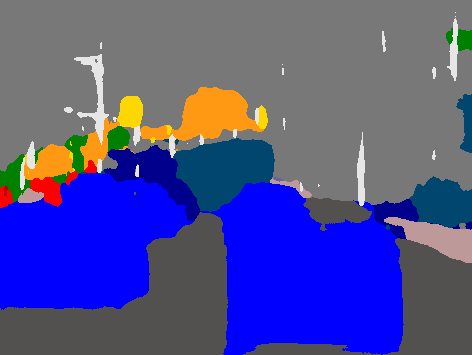}
    \\
    \raisebox{.75\height}{\rotatebox{90}{\small CBAM}} &
    \includegraphics[width=0.162\textwidth, height=0.11\textwidth]{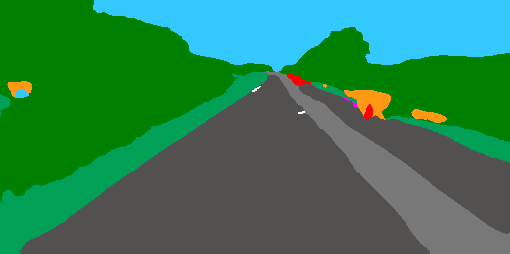} &
    \includegraphics[width=0.162\textwidth, height=0.11\textwidth]{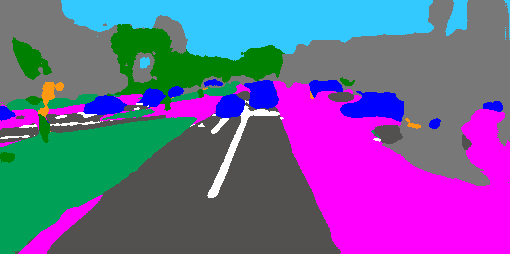} &
    \includegraphics[width=0.162\textwidth, height=0.11\textwidth]{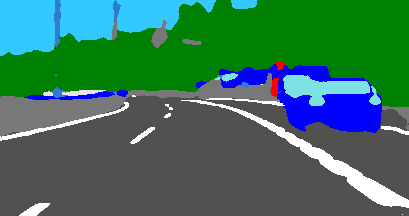} &
    \includegraphics[width=0.162\textwidth, height=0.11\textwidth]{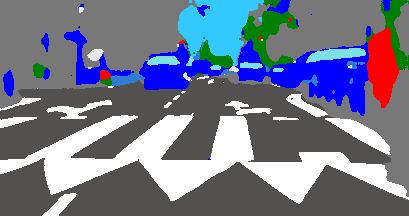} &
    \includegraphics[width=0.162\textwidth, height=0.11\textwidth]{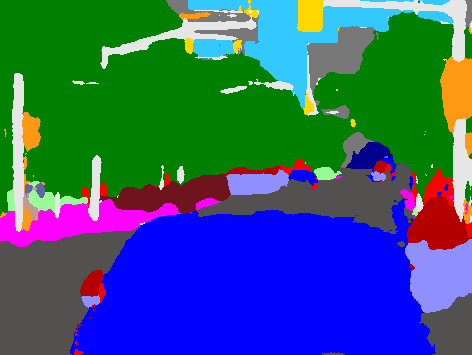} &
    \includegraphics[width=0.162\textwidth, height=0.11\textwidth]{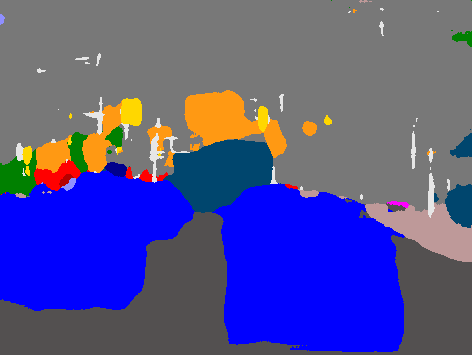}
    \\
    \raisebox{.75\height}{\rotatebox{90}{\small MSAM}} &
    \includegraphics[width=0.162\textwidth, height=0.11\textwidth]{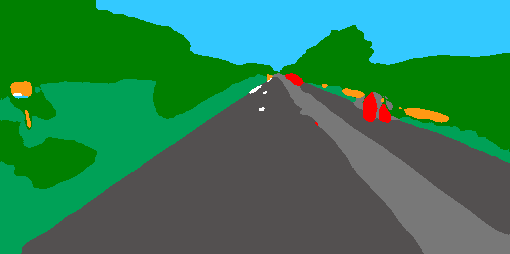} &
    \includegraphics[width=0.162\textwidth, height=0.11\textwidth]{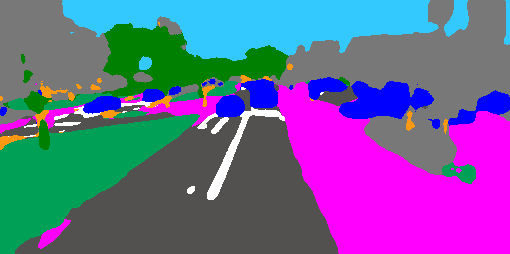} &
    \includegraphics[width=0.162\textwidth, height=0.11\textwidth]{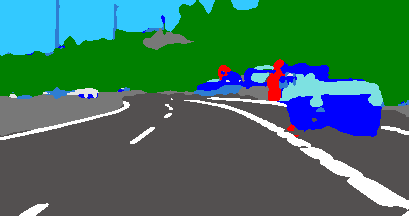} &
    \includegraphics[width=0.162\textwidth, height=0.11\textwidth]{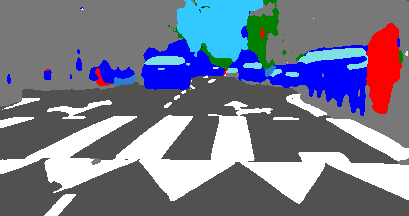} &
    \includegraphics[width=0.162\textwidth, height=0.11\textwidth]{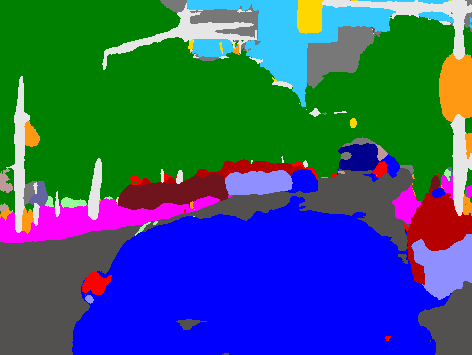} &
    \includegraphics[width=0.162\textwidth, height=0.11\textwidth]{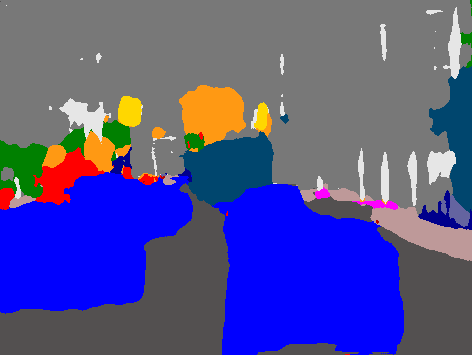}
  \end{tabular}
  \includegraphics[width=0.92\textwidth, height=0.055\textwidth]{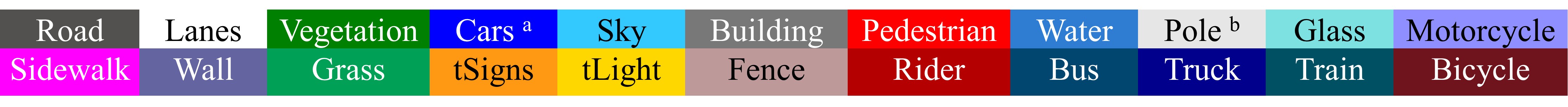}\\
  \RaggedRight
  \small\textsuperscript{a} Including Painted Metal of HSI-Drive. \quad \textsuperscript{b} Including Unpainted Metals of HSI-Drive.

  \caption{Qualitative comparison of AMs integrated into UNet$_{32}$ across HyKo-VIS (left), HSI-Drive (middle), and H-City (right). Sampled images provide relative diverse driving scenes (urban streets and highways) within the available data. MSAM variants exhibit visually consistent and competitive performance by predicting coherent object boundaries.}
\label{fig:AllSegmentationResults}
\vspace{-3ex}
\end{figure*}

\subsection{Stage~3 -- Evaluation against Other AMs}
\label{sec:Experiments - Stage3}
For comparison against established attention mechanisms, the two MSAM configurations demonstrating the most consistent top-ranking performance across both ablation datasets in Stage-2 were selected as representative candidates: MSAM$_{(1;5;11)}$ and MSAM$_{(3;7;11)}$. Table~\ref{tab:OtherBaselinesAndComputationalOverheadComparison} and Fig.~\ref{fig:AllSegmentationResults} compare them with established AMs using UNet$_{32}$ backbone on original dataset labels, averaged over three trainings:

\textbf{HyKo-VIS}: MSAM$_{(1;5;11)}$ achieves the best F1 score (80.11\%) and competitive mIoU (67.31\%) with 7.45ms GPU inference, delivering 0.45\% F1 improvement over CBAM while requiring 6.1× less GPU time (45.54ms).

\textbf{HSI-Drive}: CA achieves optimal performance (81.49\% mIoU, 90.03\% mF1), while MSAM$_{(3;7;11)}$ delivers competitive second-best results (80.70\% mIoU, 88.81\% mF1) with similar GPU efficiency (5.01 vs. 5.68ms).

\textbf{H-City}: MSAM$_{(1;5;11)}$ achieves the best performance (65.05\% mIoU, 75.45\% mF1), validating the multi-scale approach on this challenging spectral-rich urban dataset.

\textbf{Computational Overhead Analysis}: MSAM exhibits higher CPU overhead (159.01--327.50ms vs. 66.70--116.87ms) due to tensor reshaping for 1D conv. However, competitive GPU performance (5.68--9.56ms) demonstrates efficient parallelization. Consistent parameter counts (7.768M--7.801M) confirm that performance gains stem from architectural design rather than increased capacity. The competitive GPU performance and consistent accuracy on complex datasets suggest MSAM is suited for GPU-accelerated ADAS inference pipelines. However, its substantially higher CPU overhead limits direct deployment on CPU-constrained hardware without further optimisation.

\subsection{Future Directions}
Future investigations should focus on optimizing CPU bottlenecks caused by tensor manipulation limitations while preserving multi-scale attention benefits. Promising techniques include depthwise separable conv, hybrid 1D-2D Conv architectures, and channel-focused 3D Conv approaches~\cite{lin2022hybrid}. Additionally, the dataset-specific kernel selection findings (Sections~\ref{sec:Experiments}–\ref{sec:Experiments - Stage2}) highlight the need for adaptive kernel selection strategies to achieve broader applicability across diverse ADAS scenarios. These optimization efforts, combined with ongoing HSI research~\cite{gutierrez2023chip} and emerging tensor-specialized hardware accelerators, could significantly enhance MSAM's practical deployment in resource-constrained vehicular computing environments and further explore the potential of HSI in ADAS/AD applications.

\section{Conclusion}
\label{sec:Conclusion}
This paper presents a systematic investigation of multi-scale spectral feature extraction for hyperspectral segmentation in ADAS/AD through the MSAM integrated within the UNet architecture. Experimentation across all available multi-class annotated urban driving HSI datasets demonstrates the potential effectiveness of multi-scale feature extraction: MSAM integration within UNet's skip connections on average outperformed baseline UNet-SC, achieving improvements of 2.32\% in mIoU and 2.88\% in mF1 across each dataset and configurations. HyKo-VIS showed the most substantial gains (up to 6.17\% mIoU and 8.96\% mF1), demonstrating MSAM's effectiveness in handling complex spectral information. Ablation studies confirm that multi-scale kernel combinations consistently outperform single-scale approaches, with dataset-specific configurations like \textit{(1;5;11)} and \textit{(3;7;11)} showing particularly strong performance. Comparative analysis with established AMs revealed that while MSAM introduces CPU overhead due to tensor reshaping, it maintains competitive GPU latency and achieves competitive accuracy on complex HSI datasets.

The empirical findings highlight two priorities for future research: (1) adaptive kernel selection mechanisms based on dataset characteristics, and (2) architecture optimization to reduce CPU inference time while maintaining accuracy. This investigation establishes a foundation for enhanced multi-scale spectral feature extraction in automotive perception, highlighting the potential of HSI-aware spectral attention mechanisms for future ADAS/AD perception research.

\bibliographystyle{IEEEtran} 
\bibliography{references} 
\end{document}